\documentclass{llncs}
\usepackage[utf8]{inputenc}
\usepackage{graphicx}
\usepackage{algorithm, algorithmic}
\usepackage{color}
\usepackage{tabularx}

\title{Text Similarity in Vector Space Models:\\ A Comparative Study}

\author{Omid Shahmirzadi\inst{1}, Adam Lugowski\inst{2},
and Kenneth Younge\inst{1}}

\institute{TIS, EPFL, Lausanne, Switzerland\\
\email{\{omid.shahmirzadi,kenneth.younge\}@epfl.ch}
\and
Patent Research Foundation, Seattle, USA\\
\email{alugowski@patrf.org}
}

\begin{document}

\maketitle

\begin{abstract}

    Automatic measurement of semantic text similarity is an important task in natural language processing. In this paper, we evaluate the performance of different vector space models to perform this task. We address the real-world problem of modeling patent-to-patent similarity and compare TFIDF (and related extensions), topic models (e.g., latent semantic indexing), and neural models (e.g., paragraph vectors). Contrary to expectations, the added computational cost of text embedding methods is justified only when: 1) the target text is condensed; and 2) the similarity comparison is trivial. Otherwise, TFIDF performs surprisingly well in other cases: in particular for longer and more technical texts or for making finer-grained distinctions between nearest neighbors. Unexpectedly, extensions to the TFIDF method, such as adding noun phrases or calculating term weights incrementally, were not helpful in our context.
    
    \keywords{text similarity, vector space model, text embedding, patent, big data}

\end{abstract}

\section{Introduction} \label{Introduction}

Automatic detection of semantic text similarity between documents plays an important role in many natural language processing applications. Techniques for this task fall into two broad categories: structure based and structure agnostic. In the first category, solutions rely on a logical structure of the text and transform it into an intermediate structure, such as aligning trees, to do the comparison. While useful in many contexts, it often is not clear which structure to use for a particular comparison. In the second category, structure is ignored and the text is represented using a vector space model (VSM). While VSMs often do not capture semantic components of text (e.g., negations), they nevertheless have been shown to be able to measure text similarity in many applications.

A vector space model converts text into a numeric vector. A key aspect of VSMs is the definition and number of dimensions for each vector. In a common and simple approach, TFIDF defines a space where each term in the vocabulary is represented by a separate and orthogonal dimension. TFIDF measures the term frequency of each term in a text and multiplies it by the logged inverse document frequency of that term across the entire corpus. Despite its simplicity, TFIDF may suffer from an ignorance of n-gram phrases, complications with incremental updates upon addition of new documents, and a large number of dimensions. To deal with such issues, variants of TFIDF have been proposed to incorporate n-grams as new terms, and/or to adjust for the timing of the use of vocabulary across the time line of the corpus. 

Other techniques, known as text embedding, attempt to address the high-number of dimensions and the loss of semantic information in TFIDF models, by transforming each text into a low-dimensional vector. Text embedding methods can be grouped into two categories: (i) count based methods based on bag of words (where the order of words are ignored), and  (ii) prediction based methods based on sequence of words (where the order of words is taken into account). Topic models are an example of the first approach where each document is represented as a probability distribution of how relevant that document is to a given number of topics (and thus a lower-dimensional space). Each topic is selected as a weighted average of a subset of terms and document vectors are learned from the corpus on the assumption that words with similar meanings will occur in similar documents. Neural models are an example of the second approach where word vectors are learned using a shallow neural network trained from pairs of (target word, context word), where context words are taken as words observed to surround a target word. The assumption behind neural models is that words with similar meanings tend to occur in similar contexts. Document vectors can then be created out of word vectors through an averaging strategy or by considering each document as a special context token, hence obtaining document vectors directly. Prior research suggests that topic models and neural models are fundamentally similar in that they both arrive at a representation of the document in a lower-order space \cite{levy2015improving}.

In this paper, we are interested in similarity measurement between patents. Patent-to-patent similarity can have several applications such as decision making on patent filing, predicting probability of different types of patent rejections and forecasting the innovation space. Previous studies \cite{KenJeff}, have shown that TFIDF is a powerful technique to detect patent-to-patent similarity, but the performance of other vectorization methods is unknown. We therefore compare the performance of TFIDF to other, newer methods to determine the relative performance of such methods for real world problems.

This paper continues as follows: In section \ref{Background}, we discuss the background material as well as related comparative studies on semantic text similarity. In section \ref{Data}, we introduce our data gathering, pre-processing, vectorizing and performance evaluation pipeline. In section \ref{Experimental Results}, we present our experimental results. In section \ref{Conclusion}, we conclude the paper and propose some avenues for future work.


\section{Background} \label{Background}

\subsection{Vector Space Models}

Vector space models transform text of different lengths (such as a word, sentence, paragraph, or document) into a numeric vector in order to be fed into down-stream applications (such as similarity detection or machine learning algorithms). TFIDF, the most basic text vectorization method, defines a space where each term in the vocabulary is  represented  by  a  separate  and  orthogonal  dimension.  Despite its popularity and simplicity, basic TFIDF  may  suffer  from  (i) ignorance  of  n-gram  phrases,  (ii) complications  with incremental  updates  upon addition  of new  documents, and (iii) a large  number of dimensions. To deal with the two former issues, variants of TFIDF have been proposed: (i) to incorporate n-grams as new terms, and/or (ii) to adjust for the timing of the use of vocabulary across the time line of the corpus. To deal with the latter issue, text embedding methods attempt to address the high-number of dimensions by transforming each text into a low-dimensional vector. Text embedding techniques can be categorized into count-based and prediction-based models. Count-based models (a.k.a. topic models) create a document-term matrix where the weight of each cell is based on the number of times a term appears in the focal document. Prediction-based models (a.k.a. neural models) predict the occurrence of a term/document based on surrounding terms to learn a vectorization for each term/document.


In this section, we review mentioned families of the vector space models, namely (i) TFIDF models, (ii) topic models and (iii) neural models. From each family, we also select candidate methods to be compared for patent-to-patent similarity detection task.

\subsubsection{TFIDF Models.}

Term Frequency–Inverse Document Frequency (TFIDF) is one of the most common vectorization techniques for textual data with many possible variations \cite{Ramos1999}. TFIDF considers two documents as similar if they share rare, but informative, words. In TFIDF, every term is considered as a different dimension orthogonal to all other dimensions. Each term is represented by a weight which is positively correlated with its occurrence in the current document and negatively correlated with its occurrence in all other documents in the corpus. The logic behind TFIDF is to downgrade the importance of terms that are common in many documents, on the view that those terms carrying less information specific to a focal text. One common weighting scheme for a term $t$ in document $d$ is given in formula \ref{TFIDF-formula} ($|D|$ represents total number of documents in corpus, $\mbox{TF}_{t,D}$ represents total number of occurrences of term $t$ in document $d$ and $\mbox{DF}_{t,D}$ represents total number of documents in which term $t$ occurs).

\begin{equation}\label{TFIDF-formula}
\mbox{TFIDF}_{t,d} = \mbox{TF}_{t,d} \cdot \log \frac{|D|+1}{\mbox{DF}_{t,D} + 1}
\end{equation}

Despite the popularity and applicability of TFIDF to many applications, it suffers from the curse of dimensionality in many downstream applications (e.g. computing k nearest neighbors\cite{Liu04aninvestigation}), it ignores n-gram phrases, and all IDF weights might need to be updated upon the addition of new documents. The basic model, however, can be extended in several ways to avoid some of these pitfalls. We consider two recently proposed extensions in this study. 

First, we consider adding certain n-grams to the term vocabulary. N-grams allow for the combination of terms into higher-level concepts, which may be particularly important for research in computational social sciences including patent research \cite{Andersson:2016:TRN:2983323.2983858}. Adding n-grams blindly, however, would vastly increase the size of vocabulary, and thus the number of vector dimensions. A more manageable approach, therefore, is to add noun phrases based on synthetic properties of the text. We test the phrase extraction technique from \cite{Handler2016BagOW} which extracts noun phrases based on a pattern based method. They extend the simple noun phrase grammar of formula \ref{basic-noun-phrase} to support better coordination of noun phrases and better handling of textual tags. A finite state transducer is used to extract text portions that match the input grammar, including nested and overlapping parts, from the input text which is marked by part of speech (POS) tags. They impose no upper bound for the size of extracted phrases and show that their method extract high quality noun phrases efficiently.

\begin{equation}\label{basic-noun-phrase}
\mbox{Noun Phrase} \simeq (\mbox{Adj.} \mid \mbox{Noun}) \ast \mbox{Noun} (\mbox{Prep.Det.} \ast (\mbox{Adj.} \mid \mbox{Noun}) \ast \mbox{Noun}) \ast
\end{equation}

A second, and separate, extension takes advantage of the timing information of patents to implement incremental IDF \cite{Seru2017}. More specifically, whenever a new document is added to the corpus, the corresponding IDF at that point in time is calculated based on the current state of the total corpus (see formula \ref{inc-IDF}, where $T$ and $D_T$ are the addition time of a new document to the corpus and the available corpus at time $T$ respectively). Therefore, a term would have a low IDF when it is first introduced into the vocabulary and high differentiating power; and the IDF would attenuate over time as use of the term became more common. An example would be a niche term for an emerging technology, where the term would have a very high importance at the time of filing the patent, but the term would reduce in importance over time. As a convenient side property, incremental calculation of IDFs also avoids the need to update all TFIDF vectors upon addition of a new document to the corpus.

\begin{equation}\label{inc-IDF}
\mbox{TFIDF}_{t,d,T} = \mbox{TF}_{t,d} \cdot \log \frac{|D_T|+1}{\mbox{DF}_{t,D_T} + 1}
\end{equation}

\subsubsection{Topic Models.}  

Topic models transform a text into a fixed size vector, equal to a given number of latent topics. The vector represents the probability distribution that the focal text relates to each of the different topics. In practice, each topic is a weighted average of a subset of terms. Similar to TFIDF, topic models treat the text as a bag of words where order of words is ignored. On the down side, interpretation of each topic can be subjective and determining the right number of topics requires tuning of the model. 

Latent Semantic Indexing (LSI) \cite{Deerwester90indexingby,article-lsi} is a commonly used topic model to find low-dimension representation of words or documents. Latent Dirichlet allocation (LDA) is another popular topic model that fits a probabilistic model with a special prior to extract topics and document vectors. We choose LSI as the representative of topic models in this study, as LDA models can be hard to reproduce due to their highly probabilistic nature. Given a set of documents $d_i,d_2,...,d_n$ and a set of vocabulary words $w_i,w_2,...,w_m$, LSI builds a document-term matrix $X$ of $m.n$ dimensions, where item $x_{i,j}$ can represent the total occurrences of $w_j$ in $d_i$ (which can be the raw count, 0-1 count or TFIDF weight). To reduce the dimensions of $X$, truncated Singular Value Decomposition (SVD) can be applied in LSI as in formula \ref{svd} where $k$ is the number of topics.

\begin{equation}\label{svd}
X \approx U_{m,k}\Sigma_{k,k}V_{n,k}^T
\end{equation}

The low-dimensional vector of document $i$ can be obtained using $\Sigma_{k,k}\hat{d_i}$, where $\hat{d_i}$ is the $i_{th}$ row of matrix $V$. The approximation of $X$ is due to selecting the k highest items in the primary diagonal matrix $\sigma$ and corresponding columns and rows in $U$ and $V$ matrices from the original singular value decomposition. Truncated SVD can be implemented efficiently and updated incrementally on the addition of new documents \cite{10.1007/3-540-47969-4_47,Sarwar02incrementalsingular}.

\subsubsection{Neural Models.}  

Unlike models that simply count terms, neural models capture information from the context of other words that surround a given word, hence taking ordering into account. The most well-known model for predicting word context is W2V (Word to Vector)\cite{journals/corr/abs-1301-3781}, where the authors propose an algorithm based on a shallow neural network of three layers to learn word vectors. Prior research has shown the W2V model to perform well with analogy and similarity relationships. Given a context window size, the W2V algorithm comes in two forms. In the first form, known as CBOW (Continuous Bag of Words), the model predicts the probability of a target word given a context word. In the second form, know as Skip-Gram, the model predicts the probability of a context word given a target word. We explain Skip-Gram mechanics in more details. Consider a corpus with a sequence of words $w_1, w_2, ..., w_T$ and a window with size of $c$, where $c$ words on the left and right side of a focal word are considered as context. The objective functions to be maximized is given in formula \ref{skipgram}.


\begin{equation}\label{skipgram}
\frac{1}{T}\sum_{t=1}^{T} \sum_{i=-c, i \neq 0}^{c} \log p(w_{t+i} | w_{t})
\end{equation}

The probabilities are defined by Softmax. Letting $u_w$ as a target word vector and $v_w$ as a context word vector, probabilities are calculated using formula \ref{softmax}.

\begin{equation}\label{softmax}
p(w_c | w_t) = \frac{exp(v^T_{w_c}u_{w_t})}{\sum_{w=1}^{W}exp(v^T_{w}u_{w_t})}
\end{equation}

Due to the computational cost of using Softmax as a loss function (i.e., computing the gradient has a complexity proportional to the vocabulary size $W$), two efficient alternatives to Softmax have been suggested\cite{NIPS2013_5021}: hierarchical Softmax, and negative sampling. Also note that W2V ignores frequent words with a probability of $1-\sqrt{\frac{t}{f(w_i)}}$, where $t$ is a hyper-parameter of the model that is used to for the down-sampling of frequent words, and that $f(w_i)$ is the word frequency in the corpus.

To obtain document vectors from word vectors, one can average together all word vectors. Because simple averaging will give the same weight to both important and unimportant words, one may be able to retain more information by assigning weights during averaging based on TFIDF. Alternatively, an extension of W2V, known as D2V (Document to Vector) or paragraph vectors\cite{DBLP:journals/corr/LeM14}, has been proposed to obtain document vectors directly by considering a document as a special context token that can be added to the training data, such that the model can learn token vectors and consider them as document vectors. Building on W2V, the D2V algorithm also comes in two flavors: Distributed Memory (DM) and Distributed Bag of Words (DBOW). D2V can be implemented incrementally and showed a better performance compared to the previous approaches in some similarity detection benchmarks. However, on the negative side, it has numerous hyper-parameters which should be tuned to harvest its power to the full extent. We consider the D2V model as the representative neural model for the experiments in this study. 

\subsection{Similarity Metric}

Given two vectors, one can measure similarity between them in many ways: Euclidean distance, angular separation, correlation, and others. Although there are differences between each measure, in this study we adopt cosine similarity as the measure of interest (see formula \ref{cossim}) as it is well-known and frequently-used. 

\begin{equation}\label{cossim}
\mbox{sim (A,B)} = \frac{A \cdot B}{|A||B|}
\end{equation}

\subsection{Existing Comparative Studies}

Several recent studies compare different vector space models with respect to their similarity detection power for texts. Very few of them, however, target similarity detection for longer texts (e.g. documents). In \cite{DBLP:journals/corr/LauB16} the authors compared D2V variants against an averaging W2V, as well as a probabilistic n-gram model for two similarity tasks. In the first task, the goal is to detect similarity of forum questions; the second task aims to detect similarity between pair of sentences. The authors find that D2V is superior for most cases and that training the model on a large corpora can improve their results. In \cite{Naili:2017:CSW:3141923.3142185} the authors attempt to detect similarity between sentences and compare several neural models against a baseline that simply averages together word vectors. They find that more complex neural models work best for in-domain scenarios (where both the training data set and the testing data set are from the same domain), while a baseline of averaging word vectors is hard to beat for out-of-domain cases. In \cite{Arora2017ASB} the authors proposed a method for sentence embedding through the weighted averaging of word vectors as transformed by a dimensionality reduction technique. They show that their text vectors outperform well-known methods to detect sentence to sentence similarity. In \cite{DBLP:journals/corr/PagliardiniGJ17} the authors use an unsupervised method to vectorize sentences and show that their method outperforms other state-of-the-art techniques to detect similarity of short sequences of words. In each of these studies, however, the objective was to determine the performance of similarity detection algorithms on relatively short sections of text.

There has been much less research on the performance of similarity comparisons for longer text. In \cite{DBLP:journals/corr/DaiOL15} the authors compared D2V to a weighted W2V, LDA, and TFIDF to detect the similarity of documents in Wikipedia and arXiv corpus. They find that D2V can outperform other models on Wikipedia, but that D2V could barely beat a simple TFIDF baseline on arXiv. In \cite{Alvarez2017} the authors compared several algorithms to detect similarity between biomedical papers of PubMed and find that advanced embedding techniques again can hardly beat simpler baselines such as TFIDF. This paper adds to the stream of research comparing text vectorization methods for longer text. In particular, we focus on a real-world problem with an objective standard for determining similarity, whereas prior research has had to rely on broad categorizations from repositories such as Wikipedia and arXiv. To the best of our knowledge, this work is also the first comparative study of semantic similarity methods in the patent space. 

\section{Data Pipeline} \label{Data}

The Patent Research Foundation (https://www.patrf.org) provided us with a corpus of all publicly available patents from the United States Patent and Trademark Office (USPTO) from January 1976 to January 2018.\footnote{All of the necessary data can also be downloaded from https://bulkdata.uspto.gov/} From the raw data, we create a pipeline to extract the technical description from each patent and vectorize the text. Our code, based on Python 3, is available on GitHub upon email request. For the purpose of this study, we focus on the following fields:

\begin{itemize}
\item Number:      unique ID for each issued patent
\item Title:       patent information of high density
\item Abstract:    patent information of moderate density
\item Description: patent information of low density
\item Date:        date when the patent application was submitted
\item Class:       one of 491 patent main class classifications by the USPTO\footnote{https://www.uspto.gov/web/patents/classification/selectnumwithtitle.htm}
\item Subclass:    one of 82,520 patent subclass classifications by the USPTO$^4$ 
\end{itemize}

\subsection{Pre-Processing}

For pre-processing of the data, we use the DataProc service of Google Cloud Platfrom\footnote{  https://cloud.google.com/dataproc}. DataProc, a managed Apache Spark\cite{Zaharia:2010:SCC:1863103.1863113} service hosted by Google, provides a high-performance infrastructure for rapid implementation of data parallel tasks such as data pre-processing. We apply several pre-processing steps to the textual fields of patent data (title, abstract, description):

\begin{itemize}
    \item remove HTML, non-ASCII characters, terms with digits, terms less than 3 characters, internet addresses, and sequences such as DNA 
    \item stem words and change to lower-case
    \item remove stopwords, including general NLP and patent-specific ones 
    \item remove rare terms with extremely low total document frequency 
\end{itemize}

\subsection{Vectorizing}

We vectorize titles, abstracts and descriptions for each of the following models:

\subsubsection{Simple TFIDF.}

We use the machine learning library in the Apache Spark framework for our implementation of TFIDF. There are two flavors of of TFIDF in Spark to consider. The first method is based on CountVectorizer, where vocabulary is generated and term frequencies are explicitly counted before being multiplied with inverse document frequency. The CountVectorizer method, however, creates a highly sparse representation of a document over the vocabulary. The second method takes advantage of HashingTF which transforms a set of terms to fixed-length feature vectors. The hashing trick can be used to directly derive dimensional indices, but it can suffer from collisions. In this work, we use the first method as a slower but more robust technique.

\subsubsection{Incremental TFIDF.}

Incremental updating of inverse document frequencies can be implemented in one of two ways: (i) create a new TFIDF model at a fixed time interval for newly arrived patents, or (ii) create a TFIDF model on the whole corpus and then adjust document frequency vectors by subtracting document frequencies of future patents with respect to each focal patent. We implement the second approach.

\subsubsection{Phrase-augmented TFIDF.}

We augment the vocabulary of TFIDF by noun phrases as extracted from the open source NPFST library\footnote{ http://slanglab.cs.umass.edu/phrasemachine}. The library is based on Python 2, which was ported to Python 3 to be compatible with the rest of the pipeline. NPFST can be configured to limit the number of noun phrases based on their frequency and length.  

\subsubsection{LSI.}

Despite the advantage of Spark for running data parallel tasks, its support for model parallel tasks such as LSI is limited. Therefore, we use the LSI implementation in Gensim\footnote{ https://radimrehurek.com/gensim/index.html}, a well-established library for text vectorization. Gensim implements several different vector space models, exploits parallelism of multiple CPU cores, and supports large corpora that cannot be resided in the memory. Gensim also implements LSI in a memory-efficient way to support incremental updates, an important consideration for large corpora. Gensim accepts TFIDF document-word matrix as input and has several hyper-parameters, of which we consider the followings for tuning:

\begin{itemize}
    \item num-topics: number of latent topics
    \item chunksize: number of documents in each training chunk
    \item decay: weight of existing observations relative to new ones (max 1.0)
\end{itemize}

\subsubsection{D2V.}

We also use the implementation of D2V in Gensim, as the Gensim version is memory efficient, allows for incremental updates, and can be parallelized over multiple cores. We fix the random seed and Python hash seed for reproducibility of results. D2V implementation accepts raw pre-processed text in the form of TaggedDocument elements, and has several hyper-parameters, of which we consider the followings for tuning:

\begin{itemize}
    \item dm: flag to choose \textit{distributed memory} or \textit{distributed bag of words} algorithms
    \item size: dimension of the feature vectors
    \item window: maximum distance between target and context word in text
    \item sample: threshold as to which high-frequency words are randomly ignored
    \item iter: number of iterations over the corpus
    \item hs: flag to choose \textit{hierarchical softmax} or \textit{negative sampling} methods 
\end{itemize}

\subsection{Evaluation} \label{Evaluation}

To evaluate the relative performance for each vectorization method, we construct a classification test on our data, where we use the similarity score between two vectors from a given method to predict whether that pair of patents would be rated as similar (positive label) or not (negative label) according to an independent benchmark. Specifically, our evaluation test requires: \\

\begin{enumerate}
    \item a benchmark indicator of similarity as the ground truth, 
    \item a way to calculate a similarity score between vectors, 
    \item an evaluation metric to assess the accuracy of predictions, and
    \item a mechanism to tune hyper-parameters of different models.
\end{enumerate}

\subsubsection{Benchmark.}

Identifying a ground-truth benchmark for similarity is an important requirement for evaluating the performance of similarity detection from automated text vectorization. While a continuous measure of the ground truth would be ideal, we are not aware of a reliable measure of continuous patent-to-patent similarity that is separate from textual analysis. Instead, we construct a benchmark set of both \textit{more}-similar pairs of patents (positive case) as identified by the USPTO, and \textit{less}-similar pairs of patents (negative case), and evaluate the relative performance of text vectorization techniques in predicting the difference between the two cases.

For the selection of \textit{more}-similar pairs of patents, one might consider patent citations to be a natural choice -- for one would expect patent citations to reference \textit{more}-similar patents. Within the full set of patent citations, however, there is a subset of ``102 rejections" that one would expect to be even more highly-similar that an average patent citation. Patent examiners issue a 102 rejection when they believe that a cited patent is similar enough to the citing patent to cause them to reject the patent application of the citing patent on the basis that the new invention is not novel enough. Although 102 rejections are not a perfect indicator of similarity, it is reasonable to believe that 102 rejections are considerably \textit{more}-similar to each other than some other comparison sets. We therefore select 102 rejections as our human-labeled indicator of similarity.\footnote{The Patent Research Foundation provided us with pairs of patents in 102 rejections from public records for use in this study. More recently, the USPTO has also released a data set of 102 rejections\cite{rej-USPTO}. Validation tests on both data sets give similar results.}

For the selection of the comparison set of \textit{less-}similar patent pairs, one can easily identify patents that are in fact very dissimilar, and therefore very easy to distinguish from positive cases defined above. Alternatively, one can also pick pairs of patents that are, themselves, somewhat similar, and therefore much harder to distinguish from the positive cases. As such, we select and evaluate the performance of vectorization methods across a range of comparison difficulties. Going from harder separation to easier separation, we select negative cases for three testing scenarios: (i) pairs of patents selected at random from the same subclass, (ii) pairs of patents selected at random from the same main class and (iii) pairs of patents simply selected at random.

In summary, we selected a random set of 102 rejection patent pairs for our positive labels, and we selected multiple sets of patent pairs for our negative labels (sets drawn from the same subclass, the same main class, or at random). We choose the size of our test set large enough to avoid selection bias (a set of 50K positive-labeled and 50K negative-labeled pairs).

\subsubsection{Similarity Calculation.}

We calculate the similarity of each patent pair based on the cosine similarity of their corresponding vectors from a different vector space. To compare pairs of patents from an incremental TFIDF model, a recent study\cite{Seru2017} proposes replacing the IDF of the latter patent to that of the former patent so that both vectors have the same time scale to compare. Therefore we pre-compute aggregated DFs by month so that IDF replacement could be done efficiently at run time.

\subsubsection{Evaluation Metric.}

The Receiver Operating Characteristic (ROC) curve is a standard method for comparing the accuracy of classifiers in which the class distribution can be skewed and the probability threshold to assign labels is not determined\cite{Bradley:1997:UAU:1746432.1746434}. Given labels and predictions, the Area Under the Curve (AUC) represents the probability that a vectorization method will predict that a positive case (a randomly selected 102 rejection pair of patents) to be more similar than a negative case (a randomly selected pair of patents from a comparison set). The AUC is our preferred metric for comparing models, and we plot ROC curves for visual comparisons.

\subsubsection{Model Tuning.}

Models with hyper-parameters (e.g., topic models and neural models) need to be tuned for optimal performance. Studies show that hyper-parameter tuning can be as important as, or even more important than, the choice of the model itself \cite{levy2015improving}. Nevertheless, the complexity and cost of tuning can escalate quickly as the number of hyper-parameters increases. Classic approaches to tuning, such as a grid search or a random search, are often a poor choice when the cost of model evaluation is high. For grid search, it is difficult to select grid points for continuous parameters, and every added hyper-parameter will increase the tuning cost geometrically. For random search, one can end up calculating many poor configurations where model evaluation is expensive and of no use. 

Bayesian optimization is shown to be superior to classic hyper-parameter tuning solutions for expensive models \cite{Snoek:2012:PBO:2999325.2999464}. In particular, it provides a global, derivative-free approach that is suitable for tuning black box models with a high cost of evaluation. The costs of tuning is also linear with respect to the number of hyper-parameters.

Algorithm \ref{alg:bo} depicts Bayesian optimization at a high level. Given a function to optimize $f$, and a set of hyper-parameters $X$, the algorithm creates a set of initial points for the optimization surface (line \ref{op0}) and saves the obtained values in set $D$. For a budget of $N$ evaluations, the algorithm then does the following: (i) fit the distribution of possible functions as a Gaussian process $GP$ to the $D$ (line \ref{op1}); (ii) suggest the next point to assess where it maximizes the \textit{expected value} of its goodness for the probabilistic model (line \ref{op2})\footnote{This method can be extended to suggest several points instead of a single one\cite{chevalier:hal-00732512}.}; (iii) assess the costly objective function (line \ref{op3}); and (iv) add the new assessed point to the set $D$ (line \ref{op4}). At the end of the computing budget, the point with the lowest $y$ value in $D$ is reported as the optimum point.\\
\vspace{-2mm}
\begin{algorithm}

\footnotesize
\caption{Bayesian optimization loop}
\label{alg:bo}
\begin{algorithmic}
\STATE $input: f, X$
\STATE $D\gets InitialSamples(f,X)$ \label{op0}
\FOR{$i\gets |D|$ \TO $N$ } 
\STATE {$p(y|x,D)\gets FitModel(GP,D)$} \label{op1}
\STATE {$x_i\gets argmax_{x \in X}EV(x,p(y|x,D))$} \label{op2}
\STATE {$y_i\gets f(x_i)$} \label{op3}
\STATE {$D\gets D \cup (x_i,y_i)$} \label{op4}
\ENDFOR
\end{algorithmic}

\end{algorithm}
\vspace{-3mm}

We used the scikit-optimize library\footnote{ https://scikit-optimize.github.io} to implement Bayesian optimization due to the library's ability to run in parallel and to integrate with other libraries in Python. We set a tuning budget equal to 10 times the number of hyper-parameters for most experiments, but stopped tuning short for the description field due to the high cost and low expectation of further improvement.

\section{Experimental Results} \label{Experimental Results}

\subsection{Model Comparisons}

\begin{figure*} \label{exp_res}
\hspace{20mm}\scriptsize{Title} \hspace{32mm}Abstract \hspace{28mm}Description
\vspace{0mm} \\
\vspace{0mm}
 \rotatebox{90}{\scriptsize{\hspace{0.7cm}TFIDF with Phrases}}
 \includegraphics[width=.3\linewidth]{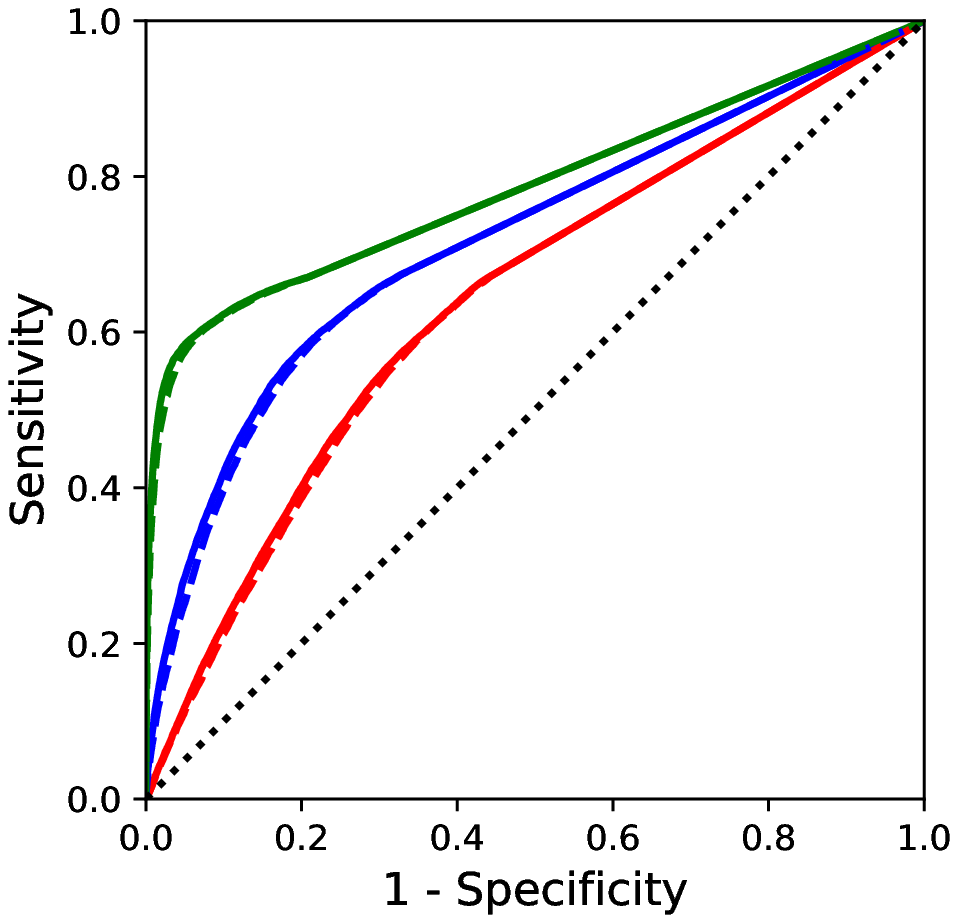} \hfill
 \includegraphics[width=.3\linewidth]{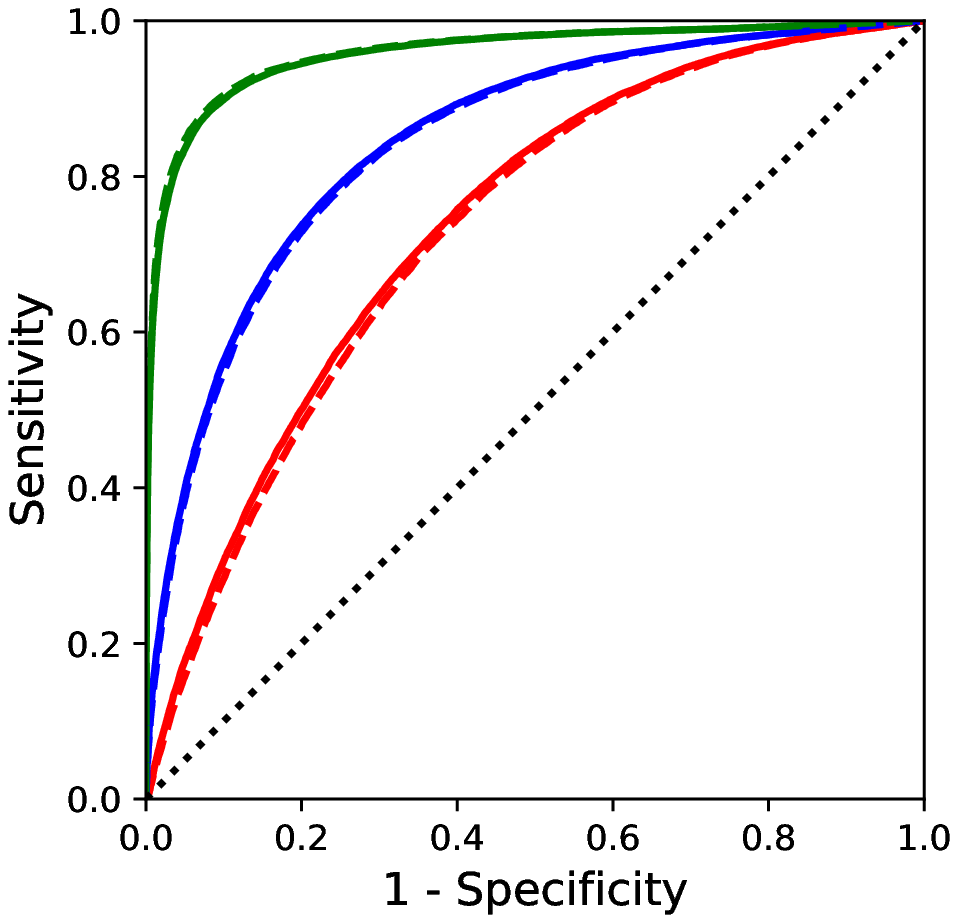} \hfill
 \includegraphics[width=.3\linewidth]{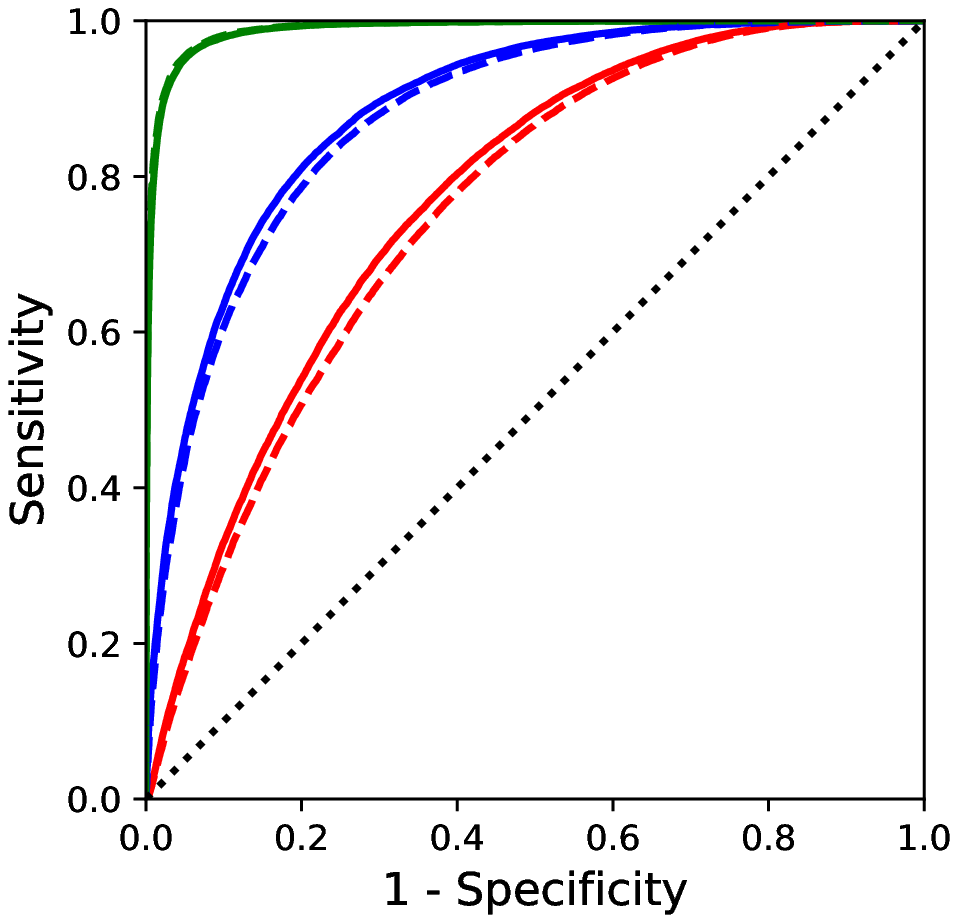} \\
 \vspace{0mm}
 \rotatebox{90}{\scriptsize{\hspace{0.8cm}Incremental TFIDF}}
 \includegraphics[width=.3\linewidth]{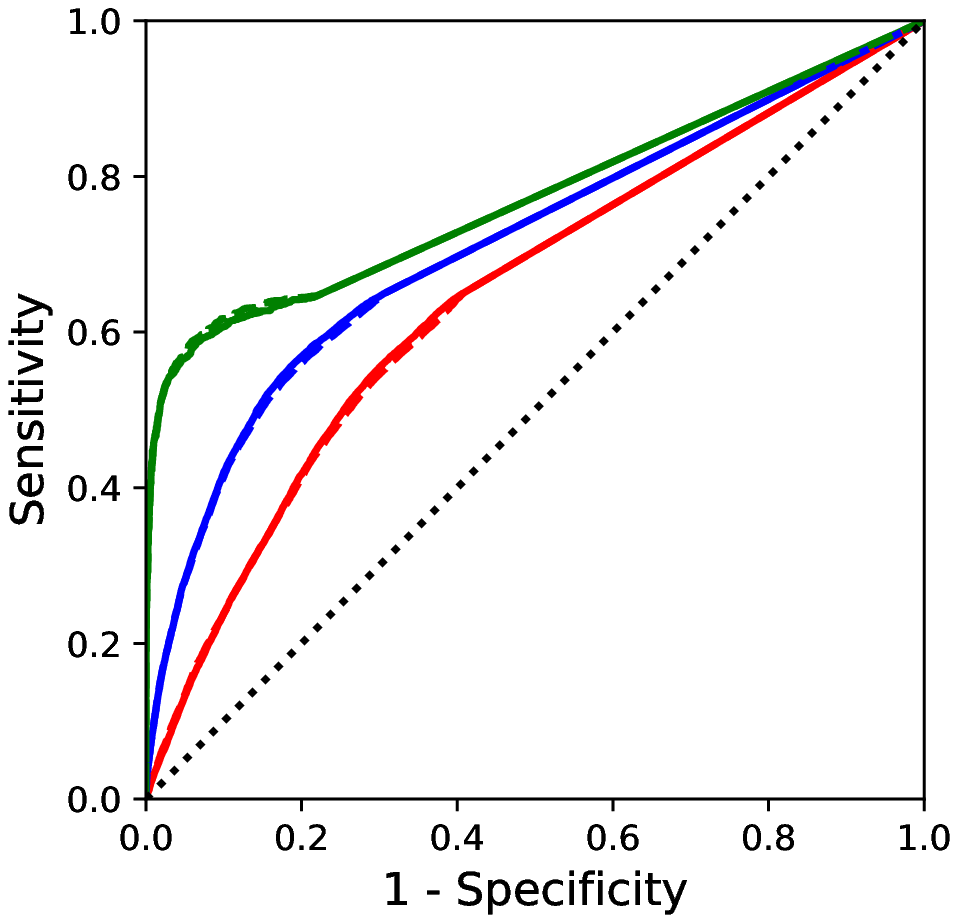} \hfill
 \includegraphics[width=.3\linewidth]{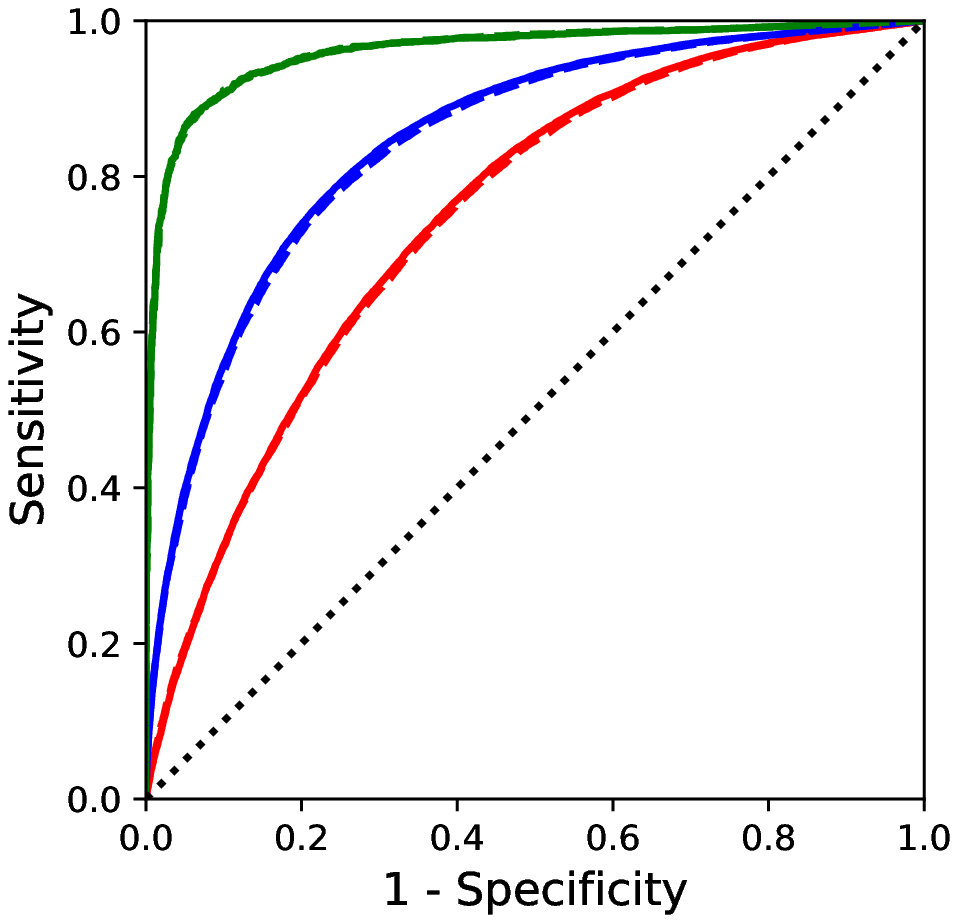} \hfill
 \includegraphics[width=.3\linewidth]{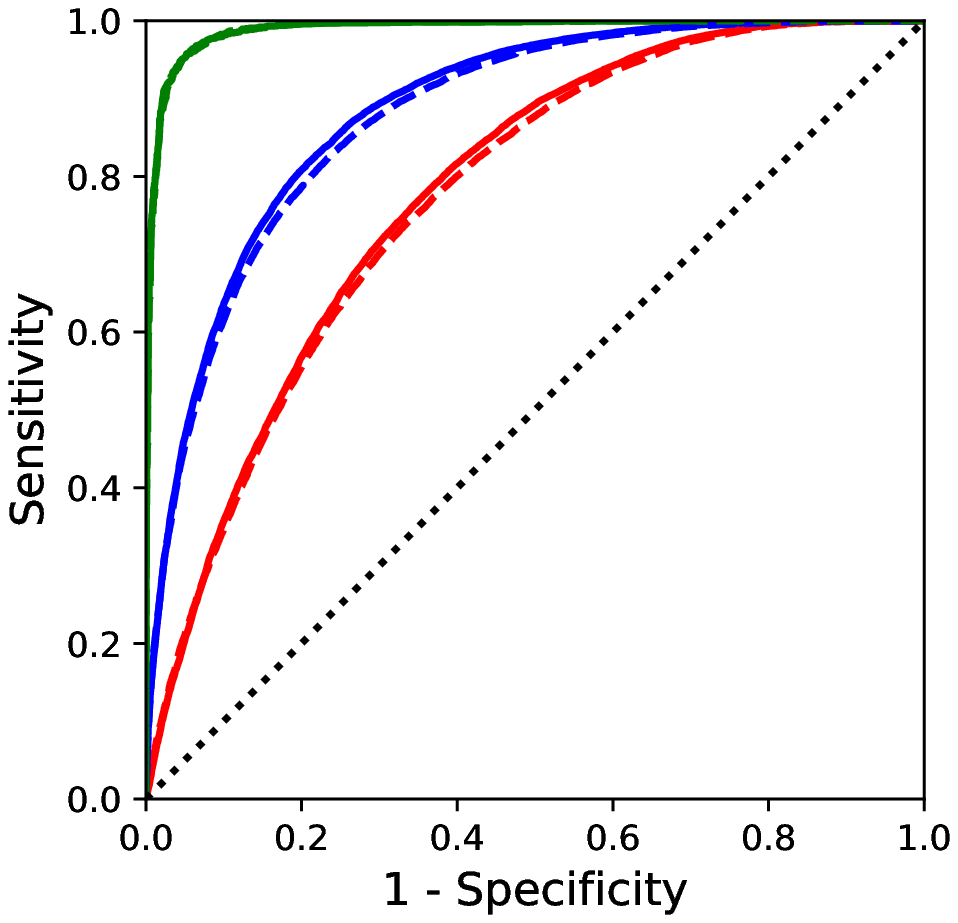} \\
 \vspace{0mm}
 \rotatebox{90}{\scriptsize{\hspace{0.9cm}Highly-Tuned LSI}}
 \includegraphics[width=.3\linewidth]{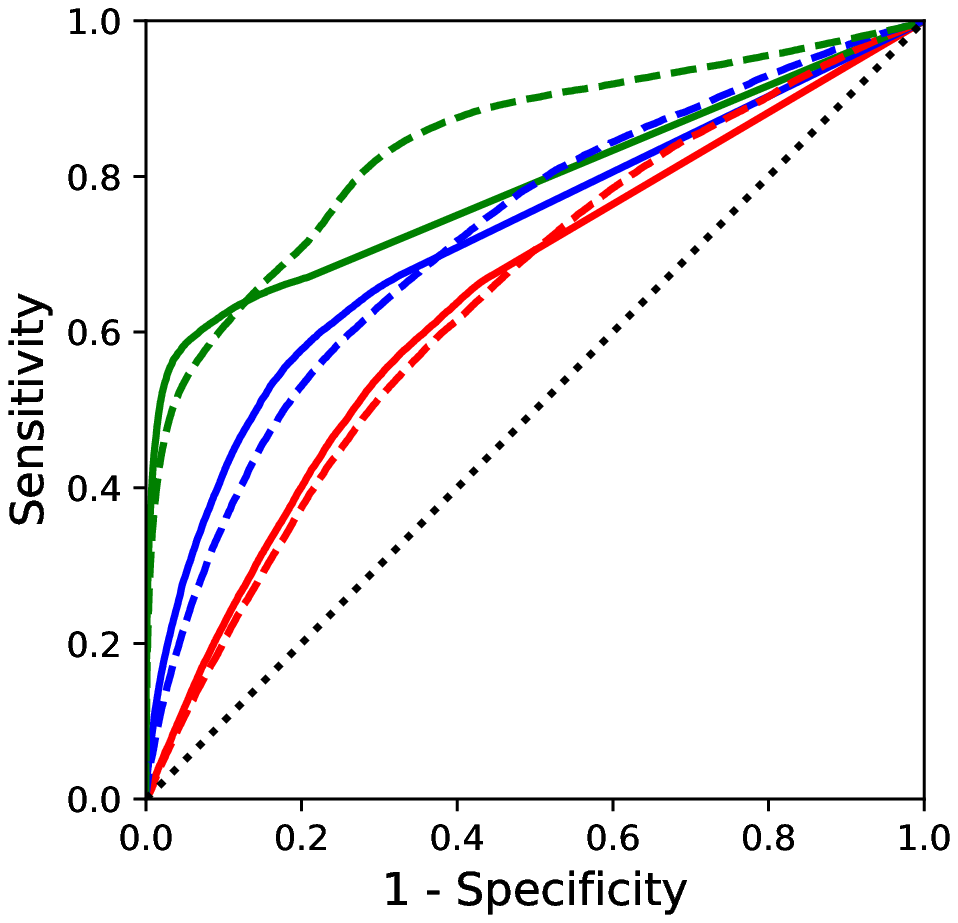} \hfill
 \includegraphics[width=.3\linewidth]{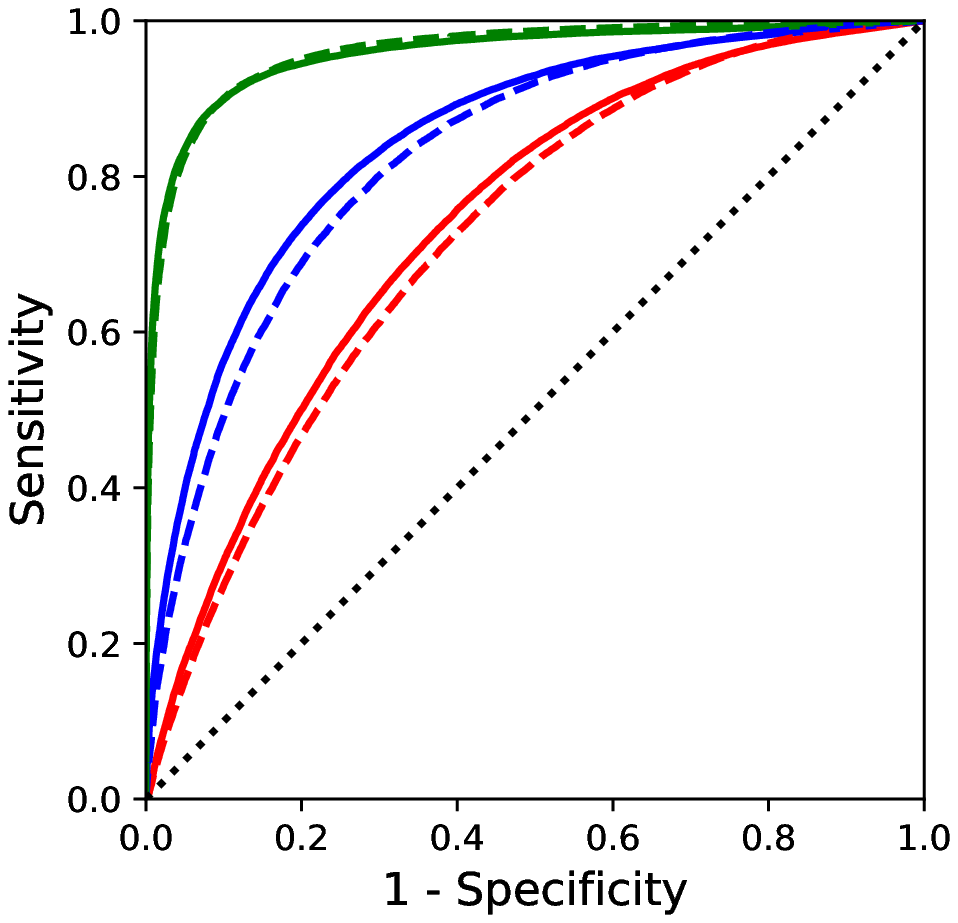} \hfill
 \includegraphics[width=.3\linewidth]{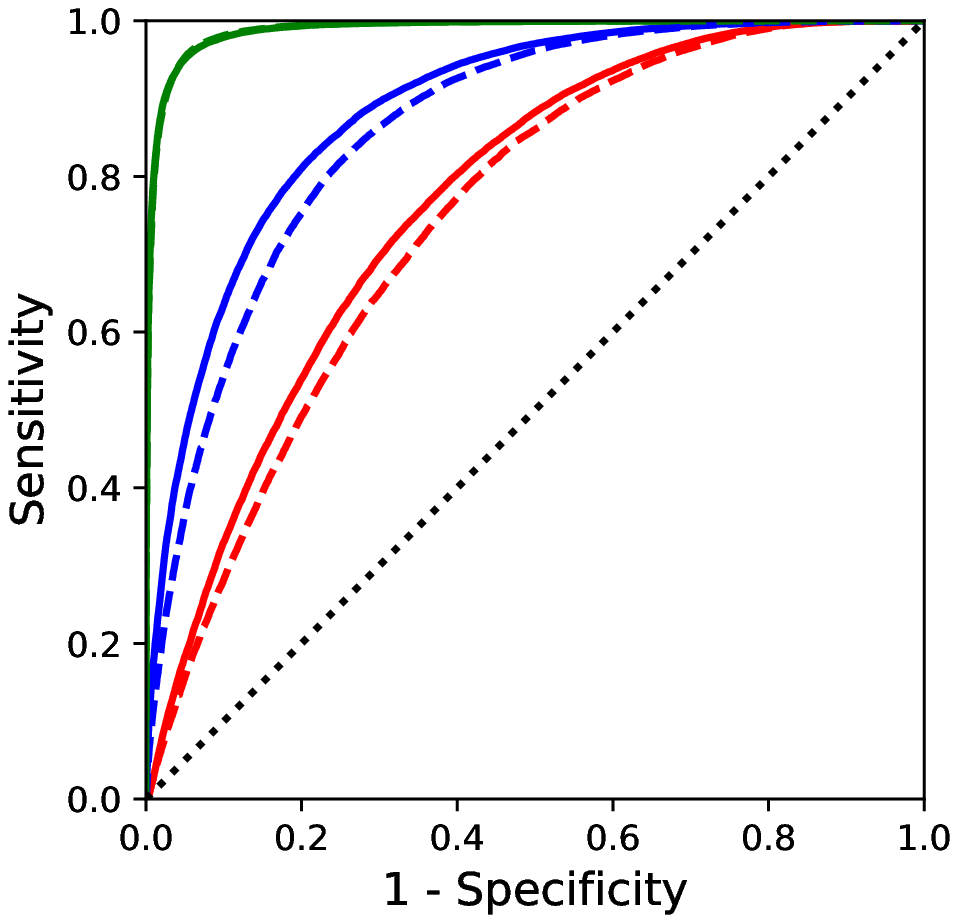} \\
 \vspace{0mm}
 \rotatebox{90}{\scriptsize{\hspace{0.9cm}Highly-Tuned D2V}}    
 \includegraphics[width=.3\linewidth]{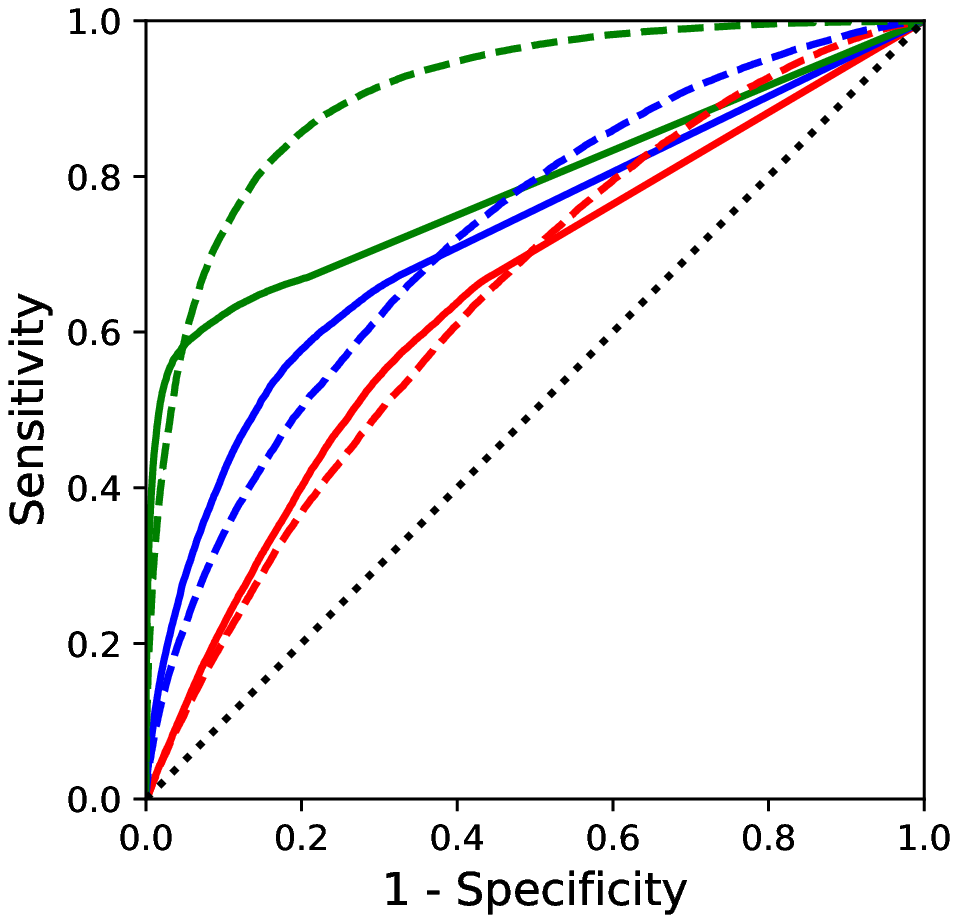} \hfill
 \includegraphics[width=.3\linewidth]{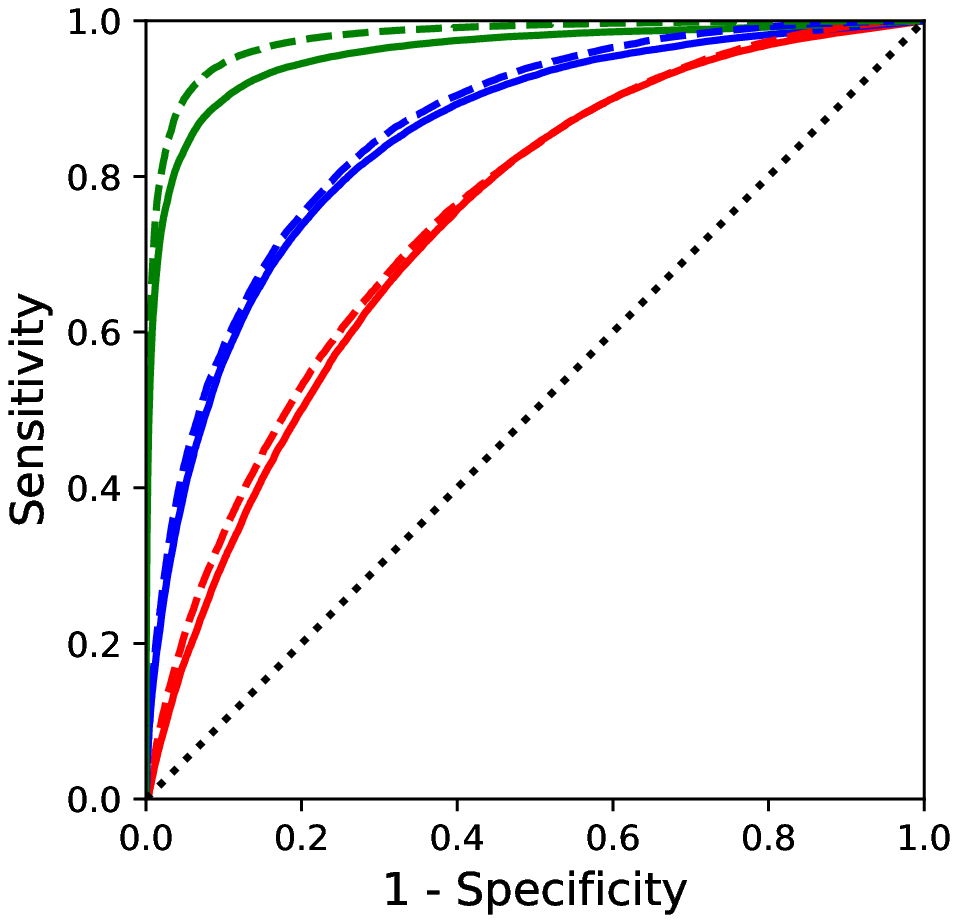} \hfill
 \includegraphics[width=.3\linewidth]{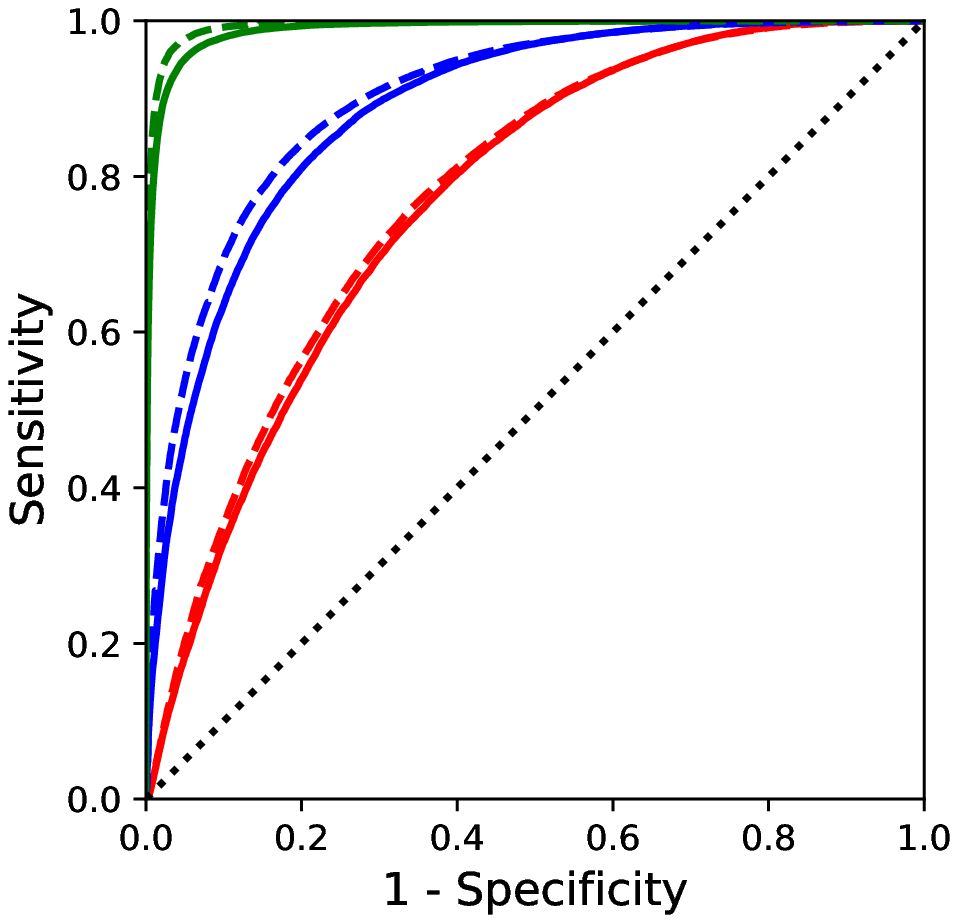} \\

 \vspace{0mm}
 \hspace{-8mm}
 \includegraphics[width=1.16\linewidth]{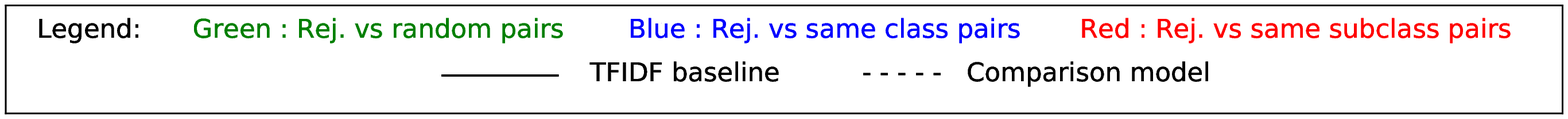}
 
\caption{Inferred accuracy of patent-to-patent similarity by vectorization method: Figures plot ROC curves for the prediction of patent rejection based on similarity score of four vectorization methods: simple TFIDF; TFIDF with noun phrases; TFIDF with incremental calculation of IDF; highly-tuned LSI; and highly-tuned D2V. In each plot, solid lines represent the simple TFIDF model as a baseline, and dashed lines represent the comparison model. Experiments were run by length of text (title, abstract and description) and difficulty of prediction: easy (in green: discrimination between a 102 rejection pair of patents and a random pair of patents); medium (in blue: discrimination between a 102 rejection pair of patents and a pair of patents from the same patent class); and difficult (in red: discrimination between a 102 rejection pair and a pair of patents from the same patent subclass).\label{exp_res}}
\end{figure*}

Figure \ref{exp_res} plots the performance of patent-to-patent similarity measurement by model. Each row corresponds to a different vectorization approach and each column corresponds to testing on a different text field. Each pane compares a simple TFIDF model as a baseline (solid lines) with a more complicated vectorization method indicated for that row (dashed lines). The comparison is done for each of the three benchmarks mentioned earlier in Section \ref{Evaluation}: a random set of 102 rejection patent pairs with positive labels, and three sets of patent pairs with negative labels (sets drawn from the same subclass, the same main class, or at random). Cosine similarity of each patent pair is calculated using the two vector space models and their corresponding ROC curves are plotted, where different colors correspond to different benchmarks.


In the first row of Fig. \ref{exp_res}, TFIDF with noun phrases performs almost identically to the baseline, across all text lengths and benchmarks. This is counter-intuitive, as augmenting a bag-of-words vector with n-grams would seem to add more information. It is possible, however, that adding \textit{every} noun phrases is too granular. We performed additional experiments with including only the top 50, 100, and 200 noun phrases, but found even worse results than the baseline. 

In the second row of Fig. \ref{exp_res}, incremental TFIDF also performs almost identically to the baseline, across all text lengths and benchmarks. This result is contrary to recent studies \cite{Seru2017} where incremental TFIDF was presumed to be beneficial. However, even if incremental TFIDF is \textit{not} better at similarity detection, it is computationally cheaper for a corpus that expands over time, and therefore our results suggest that incremental TFIDF may be a reasonable choice in that context. 

In the third row of Fig. \ref{exp_res}, a highly-tuned LSI model is able to beat the performance of the baseline, but only for a very easy similarity comparison and only for very short text (titles). Otherwise, LSI performs worse than the baseline. 

In the fourth row of Fig. \ref{exp_res}, a highly-tuned D2V model is able to beat the baseline, but this gain is considerable only for very short text (title) and only for an easy similarity comparison. D2V gives only a very slight improvement over baseline on all other conditions. It also is important to emphasize that the very minor improvement of D2V over the simple TFIDF baseline were obtained only after very extensive and expensive tuning of the D2V model.

To summarize, Table \ref{bestcase} reports the AUC and the percentage improvement in AUC of the best vectorization method in all different \textit{text field - test set} evaluation scenarios over the simple TFIDF baseline while Table \ref{walltime} depicts the rough wall time estimate required by each method to compute the corresponding vector space representation of the full corpus using our hardware setting (excluding pre-processing time). In each case, the best method\footnote{Highly tuned D2V performs the best in all testing scenarios.} performs better than a simple TFIDF model, but the percentage of improvement is negligible in most cases other than similarity comparison based on titles with a very easy distinguishability (102 rejections pairs versus random pairs). Moreover the computation wall time of the best method is at least two orders of magnitude highers than the baseline TFIDF in all scenarios.

\begin{table*} [t]
\caption{AUC performance comparison of the best VSM to the simple TFIDF. \label{bestcase}}
\scriptsize
\begin{center}
 \begin{tabular}{|p{3cm}p{2cm}p{2cm}p{2cm}|}
 \hline
  \multicolumn{1}{|c|}{ } & \multicolumn{1}{c|}{\phantom{ttttt}Title\phantom{ttttt}} & \multicolumn{1}{c|}{\phantom{ttt}Abstract\phantom{ttt}} & \multicolumn{1}{c|}{\phantom{t}Description\phantom{t}}  \\ 
 \hline
 \multicolumn{1}{|l|}{Sub-class}&
 \multicolumn{1}{l|}{}&
 \multicolumn{1}{l|}{}&
 \multicolumn{1}{l|}{}\\
 \multicolumn{1}{|r|}{Best VSM} & \multicolumn{1}{c|}{0.646} & \multicolumn{1}{c|}{0.749} & \multicolumn{1}{c|}{0.775}\\
 \multicolumn{1}{|r|}{TFIDF} & \multicolumn{1}{c|}{0.643} & \multicolumn{1}{c|}{0.738} & \multicolumn{1}{c|}{0.768} \\
 \multicolumn{1}{|r|}{Improvement} &  \multicolumn{1}{c|}{0.4\%} &  \multicolumn{1}{c|}{1.5\%} & \multicolumn{1}{c|}{0.9\%}\\
 \hline
 \multicolumn{1}{|l|}{Main-class}&
 \multicolumn{1}{l|}{}&
 \multicolumn{1}{l|}{}&
 \multicolumn{1}{l|}{}\\
 \multicolumn{1}{|r|}{Best VSM} & \multicolumn{1}{c|}{0.723} & \multicolumn{1}{c|}{0.858} & \multicolumn{1}{c|}{0.900}\\
 \multicolumn{1}{|r|}{TFIDF} & \multicolumn{1}{c|}{0.720} & \multicolumn{1}{c|}{0.846} & \multicolumn{1}{c|}{0.886} \\
 \multicolumn{1}{|r|}{Improvement} &  \multicolumn{1}{c|}{0.4\%} &  \multicolumn{1}{c|}{1.4\%} & \multicolumn{1}{c|}{1.6\%}\\
 \hline
 \multicolumn{1}{|l|}{Random}&
 \multicolumn{1}{l|}{}&
 \multicolumn{1}{l|}{}&
 \multicolumn{1}{l|}{}\\
 \multicolumn{1}{|r|}{Best VSM} & \multicolumn{1}{c|}{0.907} & \multicolumn{1}{c|}{0.977} & \multicolumn{1}{c|}{0.993}\\
 \multicolumn{1}{|r|}{TFIDF} & \multicolumn{1}{c|}{0.786} &  \multicolumn{1}{c|}{0.957} & \multicolumn{1}{c|}{0.988} \\
 \multicolumn{1}{|r|}{Improvement} & \multicolumn{1}{c|} {15.4\%} &  \multicolumn{1}{c|}{2.0\%} & \multicolumn{1}{c|}{0.5\%}\\
 \hline
 \end{tabular}
 \vspace{-5mm}
\end{center}
 
\end{table*}

\begin{table*} [t]
\caption{Computation wall time comparison of the best VSM to the simple TFIDF. \label{walltime}}
\scriptsize
\begin{center}
 \begin{tabular}{|p{3cm}p{2cm}p{2cm}p{2cm}|}
 \hline
  \multicolumn{1}{|c|}{ } & \multicolumn{1}{c|}{\phantom{ttttt}Title\phantom{ttttt}} & \multicolumn{1}{c|}{\phantom{ttt}Abstract\phantom{ttt}} & \multicolumn{1}{c|}{\phantom{t}Description\phantom{t}}  \\ 
 \hline
 
 \multicolumn{1}{|l|}{Best VSM} & \multicolumn{1}{c|}{hours} & \multicolumn{1}{c|}{days} & \multicolumn{1}{c|}{weeks}\\
 \multicolumn{1}{|l|}{TFIDF} & \multicolumn{1}{c|}{seconds} & \multicolumn{1}{c|}{minutes} & \multicolumn{1}{c|}{hours} \\
 \hline

 \end{tabular}
 \vspace{-7mm}
\end{center}
 
\end{table*}

\subsection{Hyper-parameter Tuning}

\begin{figure} [!htb]
\vspace{-5mm}
\hspace{19mm}\scriptsize{Title} \hspace{31mm}\scriptsize{Abstract} \hspace{28mm}\scriptsize{Description}
\vspace{0mm}\\
\rotatebox{90}{\hspace{0.3cm}\scriptsize{Same Subclass}}
\hspace{-0.3mm} \includegraphics[width=.3\linewidth]{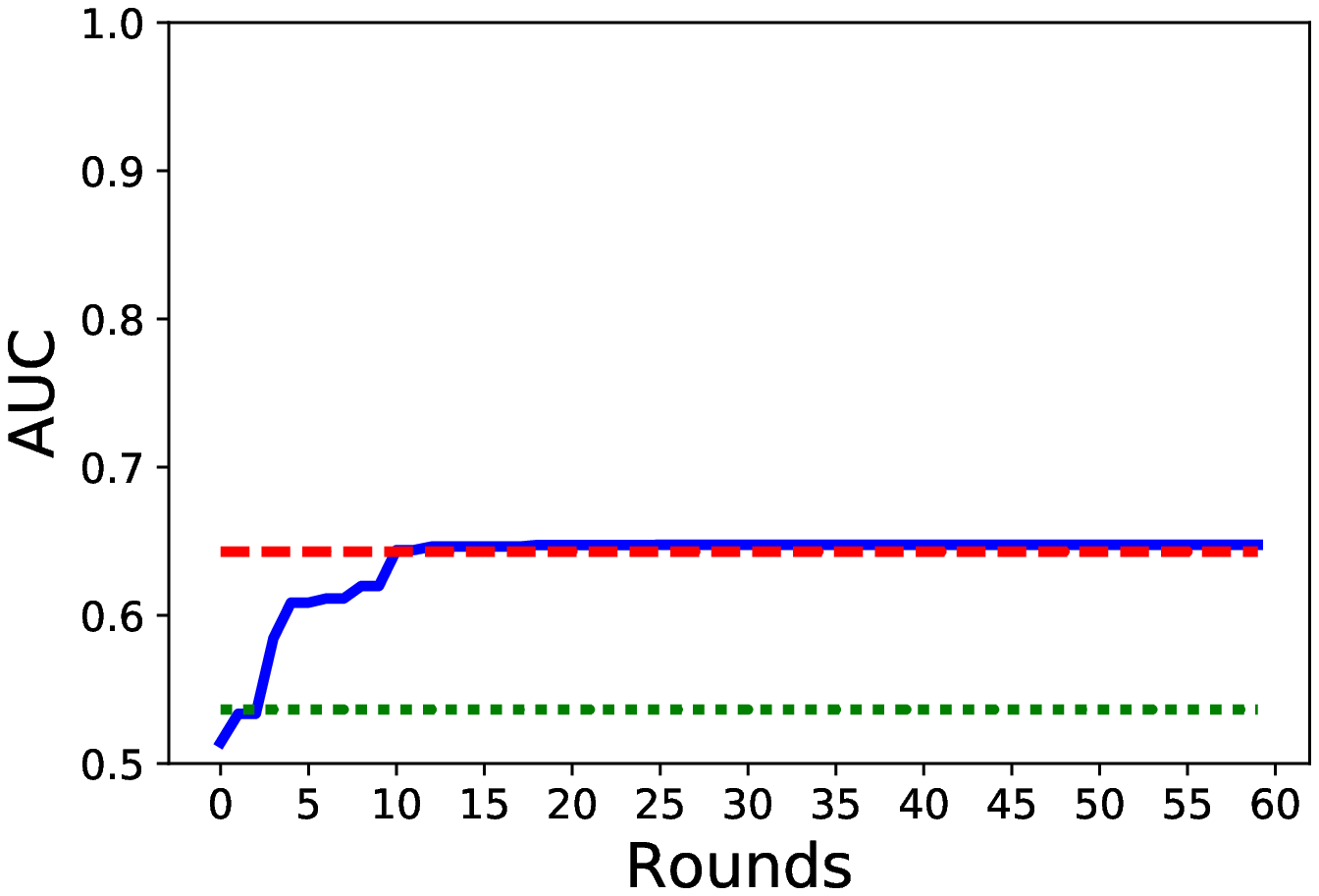} \hfill
 \includegraphics[width=.3\linewidth]{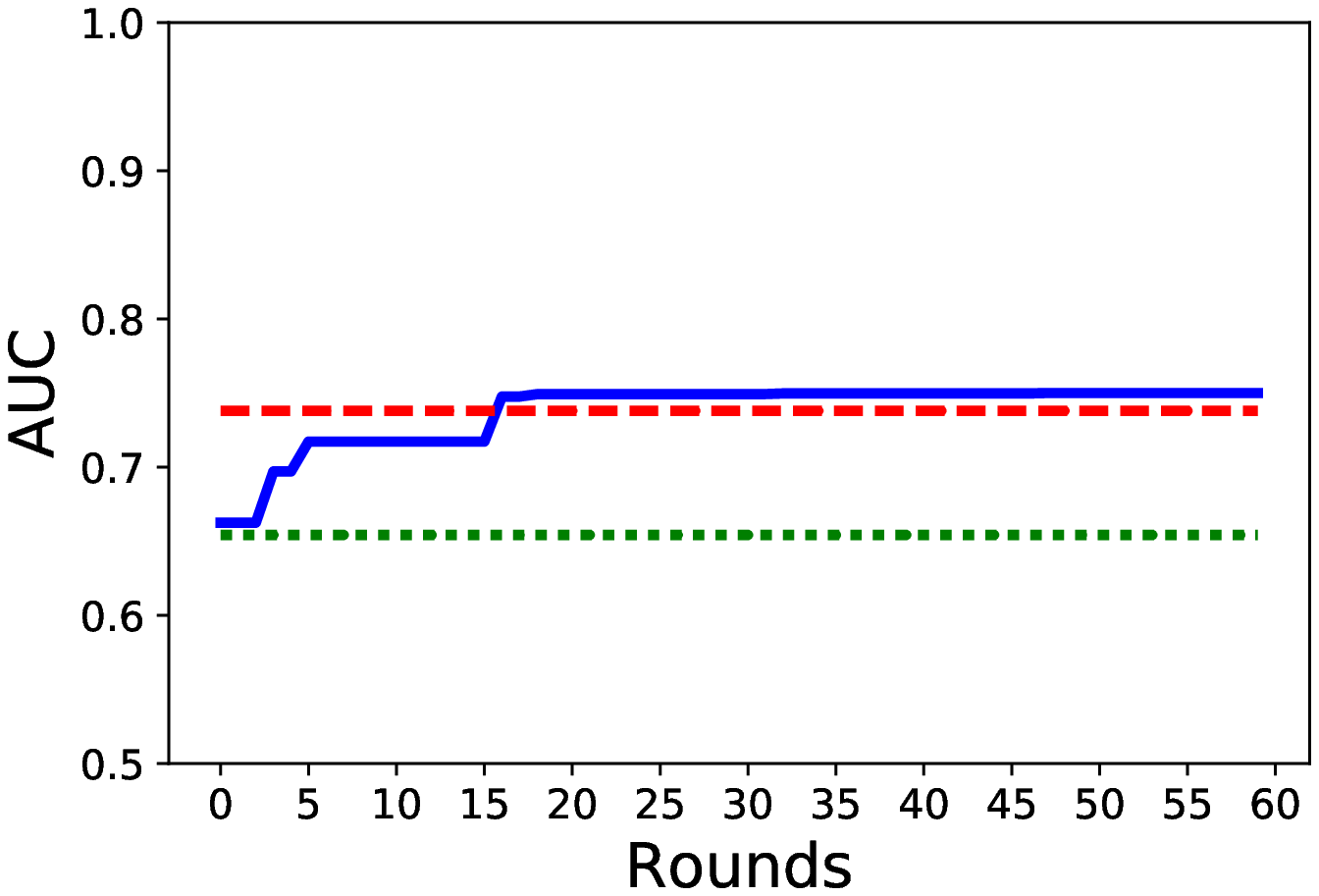} \hfill
 \includegraphics[width=.3\linewidth]{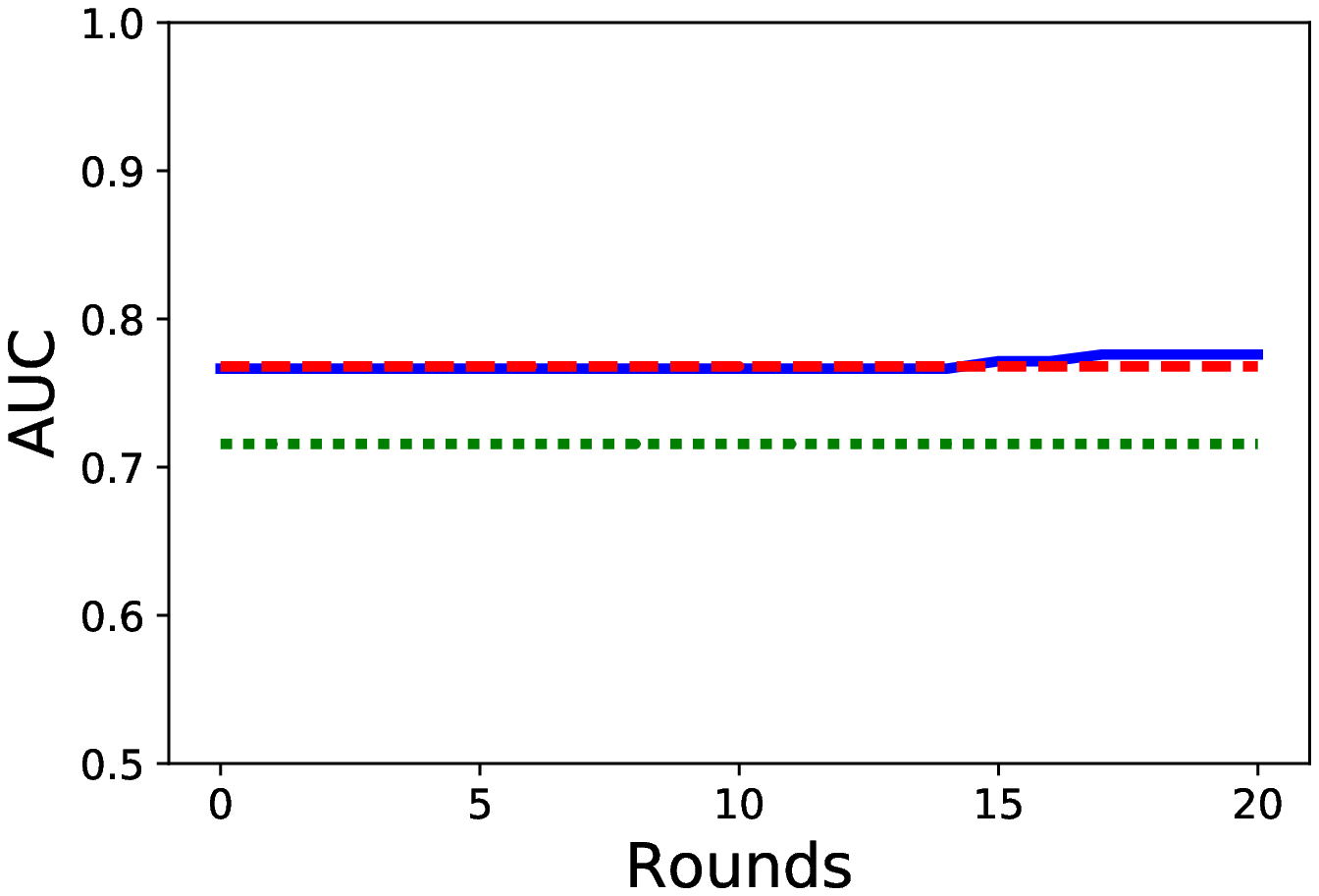} \\

 \vspace{-3mm}
 \rotatebox{90}{\hspace{0.5cm}\scriptsize{Same Class}}
 \includegraphics[width=.3\linewidth]{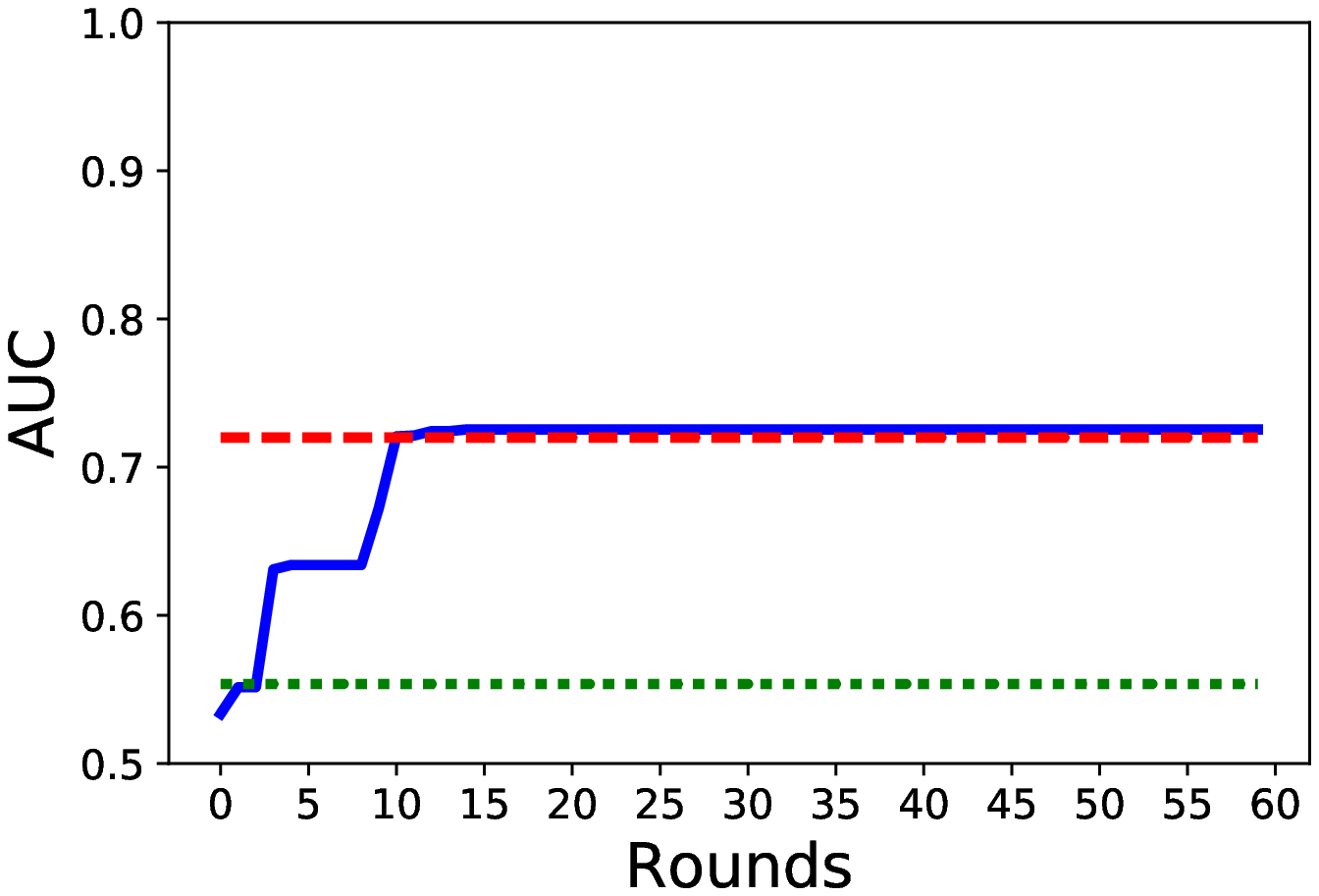} \hfill
 \includegraphics[width=.3\linewidth]{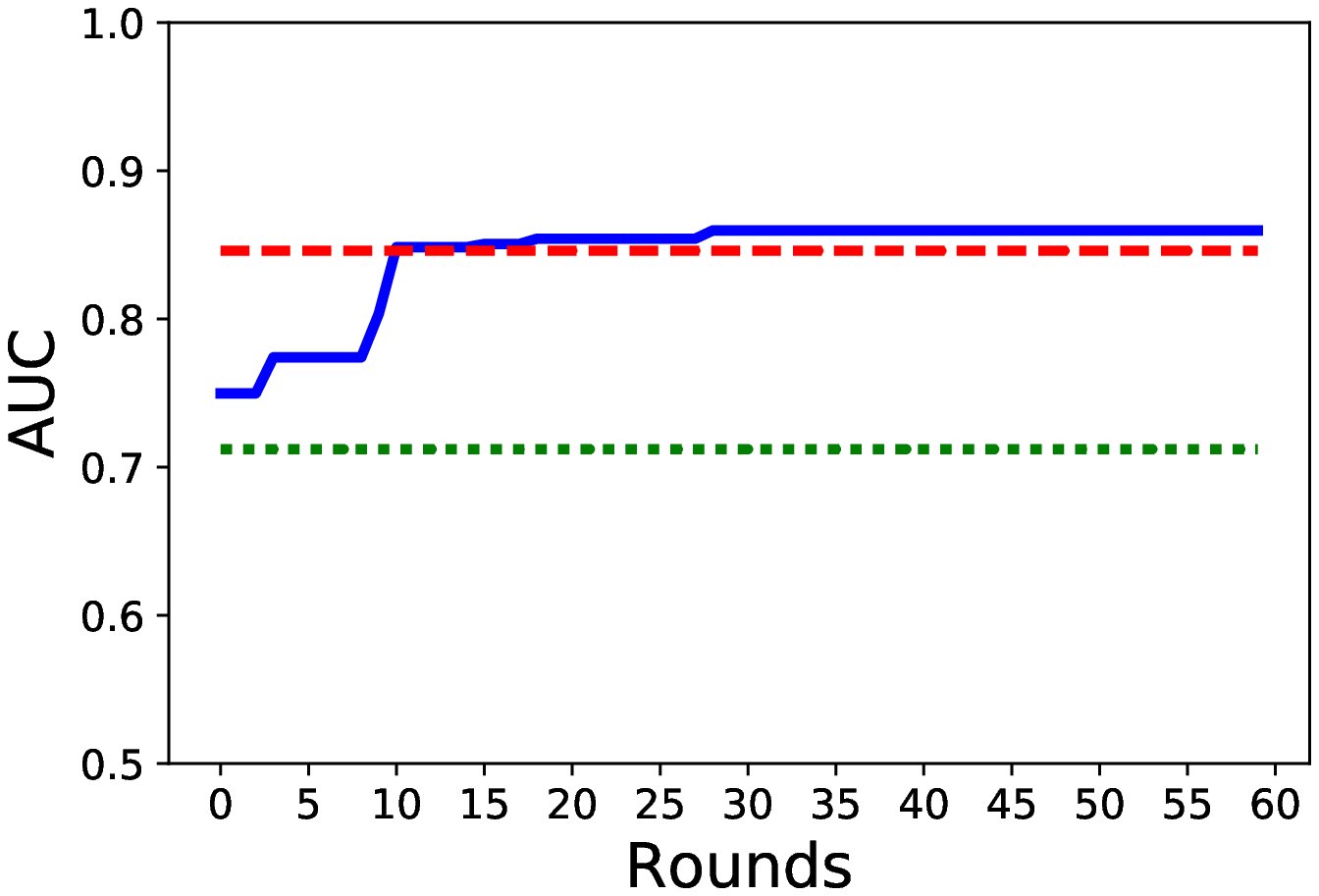} \hfill
 \includegraphics[width=.3\linewidth]{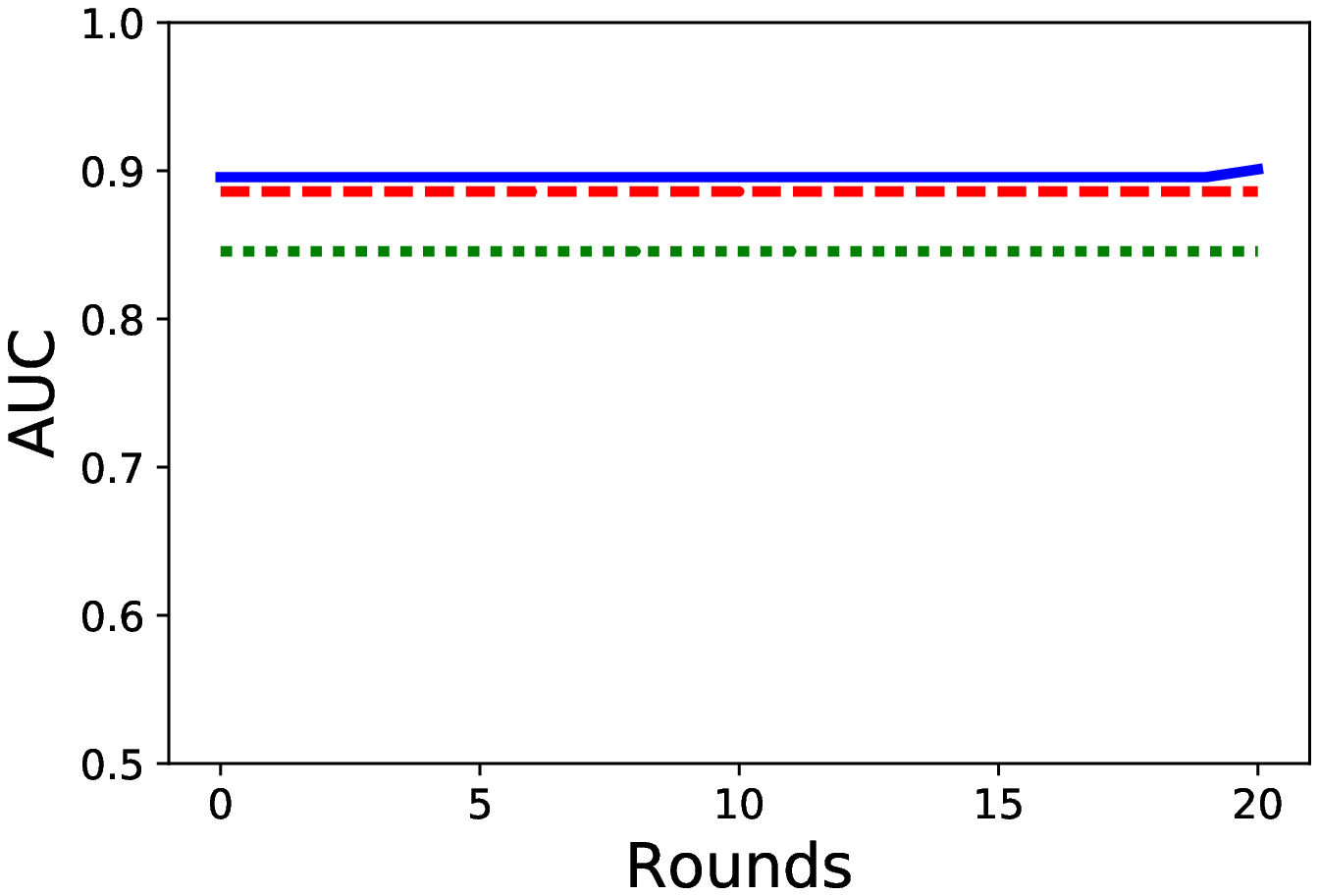} \\

 \vspace{-3mm}
 \rotatebox{90}{\hspace{0.7cm}\scriptsize{Random}}
 \includegraphics[width=.3\linewidth]{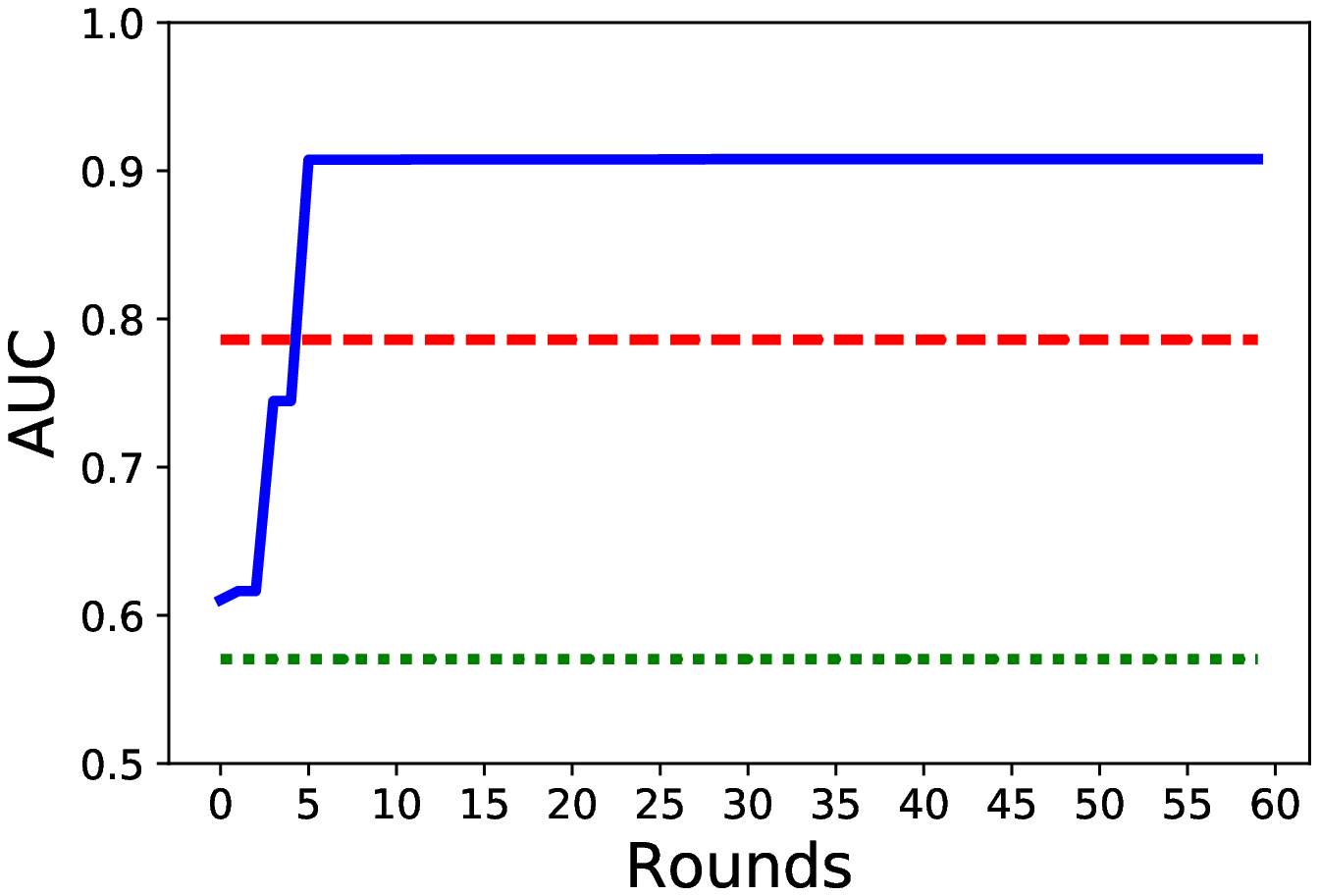} \hspace{3mm}
 \hspace{0.1mm}
 \includegraphics[width=.3\linewidth]{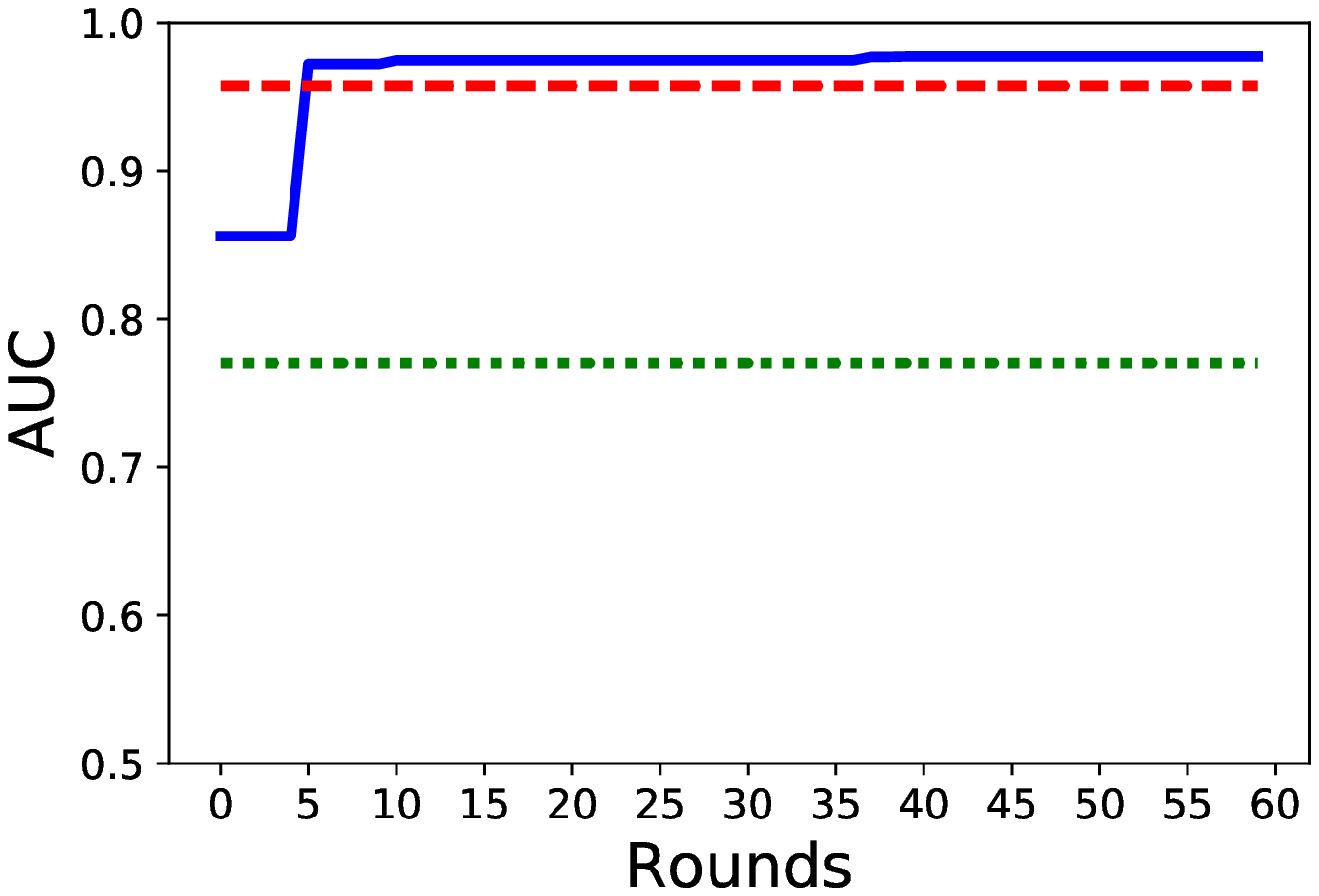} \hspace{2mm}
 \hspace{0.3mm} \includegraphics[width=.3\linewidth]{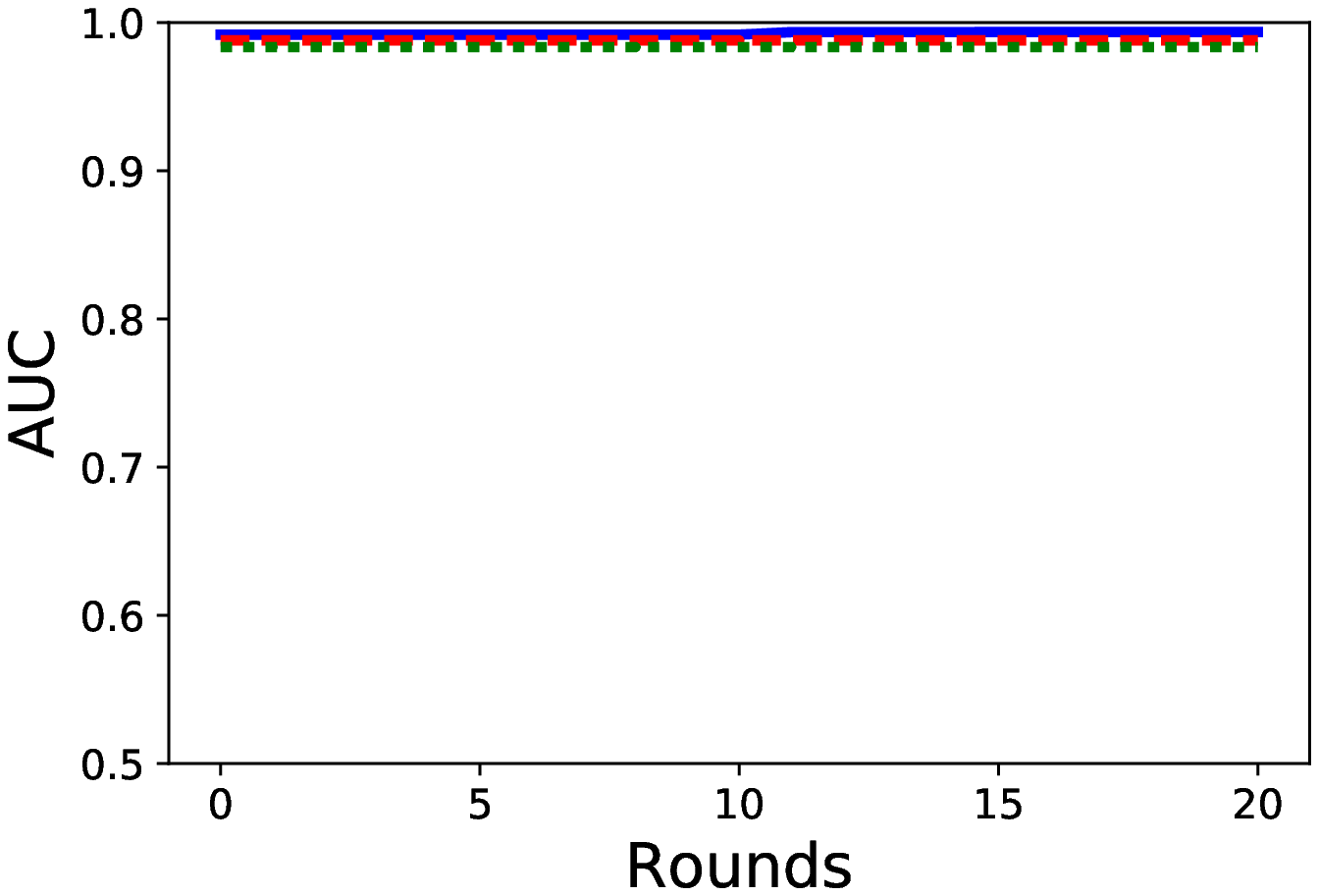}
 \vspace{0mm}
 \\
 \vspace{-1mm}
 \hspace{-12mm}
 \includegraphics[width=1.17\linewidth]{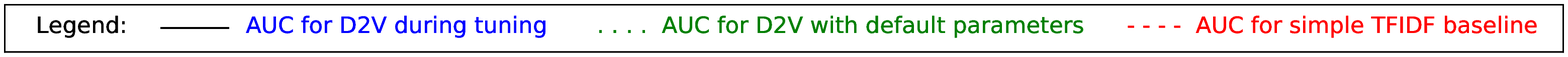} 
  
\caption{Hyper-parameter tuning of D2V for each testing scenario. \label{D2V_tuning}}  
\vspace{-3mm}
\end{figure}

To better understand the effect of tuning hyper-parameters using Bayesian Optimization, Figure \ref{D2V_tuning} plots the AUC of D2V over the course of tuning the model. Rows correspond to testing difficulty; columns correspond to text fields. We stopped tuning at 10 times the number of parameters (i.e., 60 rounds), except for Description where there was little improvement. 

Several trends in Figure \ref{D2V_tuning} are clear: 1) easier similarity detection makes hyper-parameter tuning more important; 2) longer text makes hyper-parameter tuning less beneficial; and 3) using D2V with default parameters gives very poor performance - substantially worse than a simple TFIDF baseline. Table \ref{hyperparam} shows the values of tuned hyper-parameters of D2V.

\vspace{-3mm}

\begin{table} [!htb]
\caption{Highly-tuned hyper-parameters for D2V. \label{hyperparam}}
\scriptsize
\begin{center}
 \begin{tabular}{|c|c|c|c|c|c|c|c|c|}
 
 \hline
 \phantom{ttttt}Field\phantom{ttttt} & \phantom{t}Benchmark\phantom{t} &\phantom{t} dm\phantom{t} & \phantom{t}hs\phantom{t} & \phantom{t}size\phantom{t} & \phantom{t}window\phantom{t} & \phantom{t}sample\phantom{t} & \phantom{t}iter\phantom{t} & \phantom{t}AUC\phantom{t} \\ [0.5ex] 
 \hline\hline
 Title & Sub-class & 0 & 1 & 374 & 1 & 1e-3.28 & 10 & 0.647 \\ 
 \hline
 Title & Main-class & 0 & 1 & 250 & 10 & 1e-3 & 10 & 0.725 \\
 \hline
 Title & Random & 0 & 0 & 321 & 1 & 1e-3.08 & 10 & 0.907 \\
 \hline
 Abstract & Sub-class & 1 & 1 & 491 & 1 & 1e-4.06 & 10 & 0.750 \\
 \hline
 Abstract & Main-class & 1 & 0 & 290 & 1 & 1e-4.01 & 10 & 0.859 \\
 \hline
 Abstract & Random & 1 & 0 & 522 & 1 & 1e-4.04 & 9 & 0.977 \\
 \hline
 Description & Sub-class & 1 & 0 & 321 & 7 & 1e-4.91 & 10 & 0.775 \\
 \hline
 Description & Main-class & 1 & 1 & 592 & 1 & 1e-5.52 & 10  & 0.900 \\
 \hline
 Description & Random & 1 & 0 & 501 & 1 & 1e-7 & 10  & 0.993 \\ 
 \hline
\end{tabular}
\vspace{-5mm}
\end{center}

\end{table}

\section{Conclusion} \label{Conclusion}
In this paper, we evaluated the performance of text vectorization methods for the real-world application of automatic measurement of patent-to-patent similarity. We compared a simple TFIDF baseline to more complicated methods, including extensions to the basic TFIDF model, LSI topic model, and D2V neural model. We tested models on shorter to longer text, and for easier to harder problems of similarity detection. 


For our application, we find that the simple TFIDF, considering its performance and cost, is a sensible choice. The use of more complex embedding methods which can require extensive tuning, such as LSI and D2V, is only justified if the text is very condensed and the similarity detection task is relatively coarse. Moreover extensions to the baseline TFIDF, such as adding n-grams or incremental IDFs, does not seem to be beneficial. Although our conclusion is based on experiments over patent corpus, we believe that it can be generalized to other corpora due to the minimal patent-specific interventions in our pipeline.

Our results are compatible with previous studies on semantic text similarity detection of embedding methods. The focus of prior research, however, has typically been on short text and simple similarity detection problems. Few studies have evaluated the performance of different vector space models on long text or for more challenging benchmarks. 

For the context of this study (patent-to-patent similarity) in practice, discriminating between random pairs of patents and rejection pairs of patents is a rather trivial problem that probably does not require a complicated NLP solution. It is only on such problems, however, that D2V and LSI might outperform the TFIDF model considerably. Instead, for many applications, users are often looking for the automatic detection of differences between relatively similar patents (e.g. same-subclass pairs versus rejection pairs). For such problems, where the differences in similarity are small, simple TFIDF appears to be a good choice. The difference in cost and simplicity is to the extent that the use of a simple TFIDF model, which might do slightly worse under certain conditions then more complex models, may still be justified. An extension of TFIDF with incremental IDF calculation could provide additional benefit of avoiding calculation of all TFIDF vectors upon addition of new patent(s) without scarifying performance for similarity detection task.

This study can be extended by future research in several directions, both in theory and in practice. We observed that incorporating noun phrases and incremental timing information did not lead to better detection of similar patents. For the case of noun phrases, perhaps the low weights of locally-filtered phrases misses the signal, and a more global approach for filtering might improve the performance. For the case of incremental TFIDF, it appears that adjusting IDF vectors based on time is not strong enough to affect the model performance in our context; it remains to be seen, however, how incremental IDFs may affect more rapidly evolving domains. Also studying the affect of other similarity metrics, than cosine similarity, can introduce another dimension for future work. Last, but not the least, coming up with better unsupervised vector space models for similarity detection of longer text as well as fundamental understanding of limitations of current embedding methods in this context are clearly fertile research playgrounds.

\bibliographystyle{plain}
\bibliography{references}
\end{document}